\documentclass[final,3p,times,twocolumn,authoryear]{elsarticle}

\usepackage{amssymb}
\usepackage{amsmath}
\usepackage{amsfonts}
\usepackage{graphicx}

\makeatletter
\let\c@author\relax
\makeatother


\usepackage{biblatex}
\usepackage{balance}       
\usepackage{graphicx}
\usepackage[T1]{fontenc}   
\usepackage{txfonts}
\usepackage[pdflang={en-US},pdftex]{hyperref}
\usepackage{booktabs}
\usepackage{textcomp}
\usepackage{algorithm} 
\usepackage{algorithmic} 
\usepackage{multirow} 
\usepackage{xcolor}

\usepackage{microtype}        
\usepackage{ccicons}       
\usepackage{lineno}
\usepackage{color}

\newcommand{\re}[1]{\textcolor{black}{#1}}

\journal{***}

\begin{document}

\begin{frontmatter}
	
\title{HyperNTF: A Hypergraph Regularized Nonnegative Tensor Factorization for Dimensionality Reduction}
\author[1]{Wanguang Yin}
\ead{yinwg@sustech.edu.cn}

\author[1]{Youzhi Qu}
\ead{quyz@mail.sustech.edu.cn}

\author[2]{Zhengming Ma}
\ead{issmzm@mail.sysu.edu.cn}

\author[1]{Quanying Liu\corref{cor1}}
\ead{liuqy@sustech.edu.cn}

\cortext[cor1]{Corresponding author}
\address[1]{Shenzhen Key Laboratory of Smart Healthcare Engineering, Department of Biomedical Engineering, Southern University of Science and Technology, Shenzhen, Guangdong, 518055, China}
\address[2]{School of Electronics and Information Technology, Sun Yat-sen University, Guangzhou, Guangdong, 510006, China}

\begin{abstract}
\re{Tensor decomposition is an effective tool for learning multi-way structures and heterogeneous features from high-dimensional data, such as the multi-view images and multichannel electroencephalography (EEG) signals, are often represented by tensors. However, most of tensor decomposition methods are the linear feature extraction techniques, which are unable to reveal the nonlinear structure within high-dimensional data. To address such problem, a lot of algorithms have been proposed for simultaneously performs linear and non-linear feature extraction. A representative algorithm is the Graph Regularized Non-negative Matrix Factorization (GNMF) for image clustering. However, the normal 2-order graph can only models the pairwise similarity of objects, which cannot sufficiently exploit the complex structures of samples. Thus, we propose a novel method, named Hypergraph Regularized Non-negative Tensor Factorization (HyperNTF), which utilizes hypergraph to encode the complex connections among samples and employs the factor matrix corresponding with last mode of Canonical Polyadic (CP) decomposition as low-dimensional representation. Extensive experiments on synthetic manifolds, real-world image datasets, and EEG signals, demonstrating that HyperNTF outperforms the state-of-the-art methods in terms of dimensionality reduction, clustering, and classification.}
\end{abstract}


\end{frontmatter}


\section{Introduction}

With increasingly advances in data collection and storage techniques in big data era, massive high-dimensional and heterogeneous data, also known as tensors, are generated in a wide range of real-world applications, such as time dynamic analysis ~\cite{balasubramaniam2020column, jokinen2019clustering}, latent factor analysis ~\cite{liu2020convergence, wu2019posterior, wu2020advancing}, feature selection ~\cite{liu2019embedded}, and classification ~\cite{lu2018structurally}. In comparison to the matrix factorization that convert these high-dimensional data into vectors for processing in the subsequent procedures. Tensor decomposition can naturally preserve the structural information of high-order data by retaining the multi-linear interactions, hence it can largely reduce the number of parameters to be estimated and preserve the natural structures of correlations to be learned ~\cite{hua2007face, tao2006elapsed, wang2006tensor}. As a natural result, a lot of tensor decomposition models have been developed for numerical computation. Among various of tensor decomposition models, Tucker and Canonical Polyadic (CP) are two of most widely-used models in signal processing and machine learning. The CP factorizes the input tensor into a sum of component rank-one tensors, and usually can obtain an unique solution. Unlike CP factorization, the resulting factor matrices and core tensor of Tucker decomposition are usually not unique due to the rotation of orthogonal group. To address such problem, a lot of algorithms have been proposed by incorporating an additional constraint, such as sparsity, smoothness and low rankness into the general framework of tensor decomposition models \cite{cichocki2007nonnegative, kim2007nonnegative, zhang2017low}. 

However, most of these tensor decomposition methods belong to the linear techniques for dimensionality reduction, which may fail to uncover the essential data structure that is nonlinear. Therefore, manifold learning is a good approach to learn the hidden semantics and intrinsically geometric structure within high-dimensional data. Many studies show that high-dimensional data, such as covariance of multichannel electroencephalogram (EEG) signals in biomedical engineering \cite{liu2017detecting}, hyperspectral images in remote sensing \cite{cohen2016environmental}, and gray-level video sequences in gesture recognition \cite{li2016mr}, can be thought of as a low-dimensional non-linear manifold embedded in a high-dimensional space, and the intrinsic manifold structure of these data can be learned via manifold learning~\cite{cai2007learning, cruceru2020computationally, li2016mr, liu2017detecting, zhao2022fast}. Hence, a lot of manifold learning algorithms have been developed for non-linear feature extraction. For instance, isometric mapping (ISOMAP)~\cite{tenenbaum290global}, locally linear embedding (LLE)~\cite{roweis2000nonlinear}, Laplacian Eigenmaps (LE)~\cite{belkin2003laplacian}, locality preserving projections (LPP)~\cite{lu2009regularized}, and low-rank preserving projections (LRPP)~\cite{lu2015low}, are the most well-known manifold learning techniques for dimensionality reduction. Specifically, ISOMAP uses geodesic distance induced by a neighborhood graph, controlling the loss of data information and maintaining the complex structures in low-dimensional space. However, constructing incorrect connections in the neighborhood graph will lead to the topological instability of ISOMAP. To overcome this problem, LLE attempts to preserve the local linearity of the nearest neighbors by applying the affinity approximation, where the local neighborhoods of a point on a manifold can be approximated by an affinity subspace spanned by the $ k $-nearest neighbors of that point~\cite{roweis2000nonlinear}. Unlike the idea of local linear in LLE algorithm, LE aims to preserve the local similarity of nearest neighbors by constructing the similarity matrix of Laplacian graph ~\cite{belkin2003laplacian}. Furthermore, to preserve the local neighborhood structure of the data, LPP constructs a certain affinity graph by the data and preserves the local geometry of the original data. Moreover, LRPP learns a low-rank weight matrix by projecting the data on a low-dimensional space, resulting in a low-rank projection and representation. Among these manifold learning algorithms, graph learning is one of the most popular techniques for nonlinear dimensionality reduction. The core idea of graph is to encodes the pairwise similarity among samples. However, the normal 2-order graph learning cannot effectively encodes the complex connections between samples due to its edge can only connect two vertices, hence hypergraph is candidate for revealing the complex connections among samples due to its edge can connect any number of vertices ~\cite{hu2014eigenvectors,hu2015laplacian, huang2018improved, sun2008hypergraph, zhou2007learning}. 

To the best of our knowledge, no existing methods do take into account high-order correlations among samples, and even incorporate hypergraph into the general framework of non-negative tensor factorization (NTF), utilizing the factor matrices associated with last mode of CP decomposition as the low-dimensional representation. Therefore, we propose a novel method, named Hypergraph Regularized Non-negative Tensor Factorization (HyperNTF) for dimensionality reduction and small sample data learning. The main contributions of this paper are as follow:
\re{
\begin{itemize}
  \item HyperNTF can serves to dimensionality reduction, due to its factor matrix corresponding with last mode of CP decomposition is utilized as the low-dimensional representation, thereby the storage consumption of input data can be largely reduced.
  \item HyperNTF inherits the merit of hypergraph in representing complex connections of samples, which can maximally maintain the geometric information of nearest neighborhoods in dimensionality reduction. The unfolding of synthetic 3-D manifolds (\textit{e.g.}, perforated spheres, Gaussian surfaces, doublets, and toroidal helices) show that hypergraph is better than the local similarity preserving of graph and local linear reconstructions by LLE algorithms (\textbf{Sec} \ref{manifold unfolding}).
  \item HyperNTF can effectively preserve the significant features in dimensionality reduction, which is essential for the subsequent tasks, such as clustering, classification, and pattern recognition. Experimental results in image datasets (\textit{e.g.} COIL20, ETH80, MNIST, USPS, Olivetti) demonstrate that HyperNTF outperforms all competing methods in cluster analysis (\textbf{Sec} \ref{cluster}). And results in EEG motor imagery (MI) and steady-state visual evoked potential (SSVEP) datasets show that HyperNTF has a closely alike performance with other methods for classification tasks (\textbf{Sec} \ref{classification}).
\end{itemize}
}

\begin{table} \label{tab:notation}
\setlength{\tabcolsep}{2.5pt}
\caption{List of important notations used in this paper.}
\scalebox{0.85}{
\begin{tabular}{lll} \hline
Notations                                                                                & Descriptions     \\ \hline
$\mathcal{X},X,\mathrm {x},x $                                                           & Tensor, matrix, vector, and scalar          \\
$\mathcal{X}^{(n)}\in \mathbb{R}^{L_{n}\times{L_{1}L_{2}\cdots L_{n-1}L_{n+1}\cdots M}}$ & Unfolding of a tensor on the $ n\mathrm{th} $ mode  \\
$\mathcal{G}\in \mathbb{R}^{J_{1}\times J_{2}\times \cdots \times M}$                    & The core tensor          \\
$ Z $                                                                                    & The reduced data         \\
$ M $                                                                                    & The number of training samples          \\
$U_n$, $n=1,\ldots,N$                                                                    & The factor matrices     \\ 
$ N $                                                                                    & The order (or dimension) of a tensor \\
$ \mathrm{I} $                                                                           & Identity matrix \\
$\times _{n}$		                                                                     & The tensor-matrix product\\
$\otimes$ 		                                                                         & Kronecker product\\
$\odot$ 		                                                                         & Khatri-Rao product\\
$\ast$		                                                                             & Hadamard product\\
$\div$                                                                                   & The element-wise division\\
$\left \langle \cdot,\cdot \right \rangle$		                                         & The inner product\\
$\left ( \cdot  \right ) ^T$                                                             & The transpose operation\\ \hline
\end{tabular}}
\end{table}

\section{Related Work}
In this section, we briefly review some related work, including Tucker \cite{cohen2016environmental}, Canonical Polyadic (CP) \cite{wang2011image}, manifold regularization non-negative Tucker decomposition (MR-NTD) \cite{li2016mr}, heterogeneous tensor decomposition via optimization on Multinomial manifold (HTD-Multinomial) \cite{sun2015heterogeneous}, and low-rank regularized heterogeneous tensor decomposition (LRRHTD) \cite{zhang2017low}.

\subsection{\re{Review of Tensor Decomposition}}

\textbf{Tucker:} Tucker model is one of the most well-known models for tensor analysis, which is expressed as the form of core tensor multiplied by a set of factor matrices in each mode. Given a non-negative object $ \mathcal{X}\in \mathbb{R}^{L_1\times L_2\cdots\times L_N} $, it can be written as follows:

\begin{equation}\label{key}
\mathcal{X}=\mathcal{G}\times_{1}U_{1}\times_{2}U_{2}\cdots \times _{N}U_{N}
\end{equation}
where $ \mathcal{G}\in \mathbb{R}^{J_1\times J_2\cdots\times J_N} $ is the so-called core tensor, and $ U_n\in \mathbb{R}^{L_n\times J_n} $ for $ n=1,\ldots,N $ are the factors matrices to be learned. Otherwise, we can also formulate the Tucker decomposition along the $ n\mathrm{th} $ mode unfolding as following
 
\begin{equation}\label{key}
\mathcal{X}^{(n)}=U_n\mathcal{G}^{(n)}\left ( U_N\otimes \dots\otimes U_{n+1}\otimes U_{n-1}\otimes\dots\otimes U_1 \right )^T
\end{equation}
Or, by incorporating an additional constraint, such as orthogonality, smoothness, sparseness, and low-rankness into the Tucker decomposition, emerging a lot of related works. For example, higher order orthogonal iteration and high order singular value decomposition (HOSVD) are two variants of Tucker model. In HOOI, all factor matrices are enforced orthogonality constraint, \textit{i.e.} $ U_n^TU_n=I_n$, $ n=1,\ldots,N $, while in HOSVD, both core tensor and factor matrices are enforced orthogonality constraints ~\cite{de2000multilinear}. 

\textbf{CP:} In the case that supper-diagonal elements of core tensor are nonzero, Tucker decomposition turns into the CP model, it can be formulated as follows

\begin{equation}\label{key}
\mathcal{X}=\Sigma \times_{1}U_{1}\times_{2}U_{2}\cdots \times _{N}U_{N}
\end{equation}
where $ \Sigma $ is the super-diagonal tensor. For a 3-order super-diagonal tensor (\textit{i.e.} $ \Sigma\in\mathbb{R}^{J\times J\times J }$), it can be written as:

\begin{equation}\label{3d}
\Sigma=fold_n\left ( \mathrm{I} \left ( \mathrm{I}\otimes \mathrm{I} \right ),\begin{bmatrix}
J & J & J
\end{bmatrix} \right )
\end{equation}
which contains $ J $ nonzero elements of unit one. $ fold_n (\cdot) $ is reshape a matrix into a tensor. Moreover, for the $ n\mathrm{th} $ mode unfolding of CP decomposition, it can be written as:

\begin{equation}\label{key}
\begin{aligned}
\mathcal{X}^{(n)}&=U_n\left ( U_N\odot \dots\odot U_{n+1}\odot U_{n-1}\odot\dots\odot U_1 \right )^T\\
&=U_n\left ( \odot_{i\neq n}U_i \right )^T
\end{aligned}
\end{equation}
where $ \left ( \odot_{i\neq n}U_i \right ) $ is a simplify representation for a set of Khatri-Rao products in all modes except the $ n\mathrm{th} $ mode.

Additionally, by incorporating the prior knowledge into the standard model of tensor decomposition, there emerges a lot of algorithms, including the non-negative Tucker decomposition (NTD) ~\cite{kim2007nonnegative}, manifold regularization NTD (MR-NTD) ~\cite{li2016mr}, graph-Laplacian Tucker tensor decomposition (GLTD) ~\cite{jiang2018image}, heterogeneous Tucker decomposition (HTD-Multinomial) ~\cite{sun2015heterogeneous}, and low-rank regularized heterogeneous tensor decomposition (LRRHTD) ~\cite{zhang2017low}.

\textbf{MR-NTD:} To preserve the geometric information in dimensionality reduction, MR-NTD incorporates the normal 2-order graph into the core tensors as the manifold regularization term, and simultaneously enforces non-negative constraints on the core tensors and factor matrices, facilitating a better physical interpretation for non-negative components within original data. Given a set of non-negative objects $ \left\{ \mathcal{X}^{\left ( i \right )} \right\}_{i=1}^M $, the objective of MR-NTD is to learn a set of core tensors and factor matrices as following

\begin{equation}\label{eq:MR-NTD}
\begin{aligned}
&\mathop{argmin}_{\left\{ U_n \right\}_{n=1}^N, \left\{ \mathcal{G}^{\left ( i \right )} \right\}_{i=1}^M}\sum_{i=1}^M\left\|\mathcal{X}^{\left ( i \right )}-\mathcal{G}^{\left ( i \right )}\times_1U_1\cdots\times _NU_N \right\|_F^2\\
&+\lambda \sum_{i=1}^M\sum_{j=1}^M \left\| \mathcal{G}^{\left ( i \right )} - \mathcal{G}^{\left ( j \right )} \right\|_F^2w_{ij}\\
&s.t. \: \: U_n\geq 0, 1\leq n\leq N; \mathcal{G}^{\left ( i \right )}\geq 0, 1\leq i\leq M\\
\end{aligned}
\end{equation}
where $ \left\{ U_n \right\}_{n=1}^N $ are the factor matrices for dimension reduction, and $ \left\{ \mathcal{G}^{\left ( i \right )} \right\}_{i=1}^M $ are the core tensors of low-dimensional representation, which are learned by using Multiplicative Update Rule (MUR) \cite{li2016mr}. \textcolor{black} {$ w_{ij} $ is the weight coefficient calculated from the $ k $-nearest neighboring search, for measuring the local similarity connections among data points, that is}

\begin{equation}\label{weight}
\begin{aligned}
w_{ij} = \left\{\begin{matrix}
 1, & if \:\mathcal{X}^{\left ( i \right ) }\in\mathcal{N}_k(\mathcal{X}^{\left ( j \right ) }) & or \:\mathcal{X}^{\left ( j \right ) }\in\mathcal{N}_k(\mathcal{X}^{\left ( i\right ) })\\
 0, & otherwise &
\end{matrix}\right.
\end{aligned}
\end{equation}
where $ \mathcal{N}_k{\left ( \cdot \right ) } $ is the $ k$-nearest neighbors of objects. A distinguish difference from our model is that MR-NTD treats each input data as an individual tensor, thereby it needs to learn a set of core tensors in each iteration. As the dimensions of data increase, the size of the core tensors increase exponentially, thereby the presence of the core tensors increase the computational cost and limit its ability to represent higher-dimensional data. 

\textbf{GLTD:} To achieve a better reconstruction of Tucker decomposition, graph-Laplacian Tucker tensor decomposition (GLTD) incorporates the normal 2-order graph into the factor matrix associated with last mode of Tucker decomposition ~\cite{jiang2018image}, that is

\begin{equation}\label{eq:GLTD}
\begin{aligned}
&\mathop{argmin}_{\left\{ U_n \right\}_{n=1}^{N-1}, Z, \mathcal{G}}\left\|\mathcal{X}-\mathcal{G}\times_1U_1\cdots\times_{N-1}U_{N-1} \times_NZ \right\|_F^2 \\
& + \lambda \sum_{i=1}^M\sum_{j=1}^M \left\| Z_i - Z_j \right\|_F^2w_{ij}\\
& s.t. \: \: U_n^TU_n=\mathrm{I}_n, n=1,\ldots,N-1\\
\end{aligned}
\end{equation}
Here, $ \mathrm{I}_n $ is an identity matrix, and $ Z $ is the reduced data. In comparison to MR-NTD, as the dimensions of data increase, the size of the factor matrix increases linearly, thereby its computational cost is reduced and its ability to represent high-dimensional data is increased.

\textbf{HTD-Multinomial:} Moreover, to simultaneously performs dimension reduction and achieves low-dimensional representation for input data, there emerges HTD-Multinomial algorithm, where the first $ (N-1) $-modes of factor matrices are endowed with orthogonality constraints and the last mode of factor matrix is equipped with simplex constraint ~\cite{sun2015heterogeneous}. Given the objective function of HTD-Multinomial as following

\begin{equation}\label{eq:HTD-Multinomial}
\begin{aligned}
&\mathop{argmin}_{\left\{ U_n \right\}_{n=1}^{N-1},Z, \mathcal{G}}\left\|\mathcal{X}-\mathcal{G}\times_1U_1\cdots\times_{N-1}U_{N-1}\times _NZ \right\|_F^2\\
& s.t. \: \: U_n^TU_n=\mathrm{I}_n, n=1,\ldots,N-1;\mathrm{1}Z=\mathrm{1}, Z\geq 0\\
\end{aligned}
\end{equation}
where $ \mathrm{1}Z=\mathrm{1} $ is the probability simplex constraints, modeling allocations or probability distribution of samples. In HTD-Multinomial, the first $ ( N-1 ) $-modes of factor matrices are solved by extracting the principal component of specific mode unfolding, and the last mode of factor matrix is learned by using a nonlinear optimization technique over the special matrix Multinomial manifold that equipped with Fisher information metric~\cite{sun2015heterogeneous}.

\textbf{LRRHTD:} Motivated by the HTD-Multinomial algorithm, the recently developed low-rank regularized heterogeneous tensor decomposition (LRRHTD) features to enforce the low-rank constraints on the last mode of factor matrix, for revealing the global structure of samples ~\cite{zhang2017low}. It can be written as follows: 

\begin{equation}\label{eq:LRRHTD}
\begin{aligned}
&\mathop{argmin}_{\left\{ U_n \right\}_{n=1}^N, \mathcal{G}}\left\|\mathcal{X}-\mathcal{G}\times_1U_1\cdots\times_{N-1}U_{N-1} \times_NZ \right\|_F^2 + \lambda\left\|Z \right\|_{\ast }\\
& s.t. \: \: U_n^TU_n=\mathrm{I}_n, n=1,\ldots,N-1\\
\end{aligned}
\end{equation}
where $ \left\| Z \right\|_{\ast } $ is the nuclear norm for low-rank constraint. Thus, the learning objective of LRRHTD is solved by using the augmented Lagrangian multiplier method. As a result, the storage consumption of LRRHTD is high-cost due to the reduced data (\textit{i.e.} $ Z $) is a square matrix that equals to the number of training samples.

To summarize, some prior knowledge, such as low rankness, orthogonality and non-negativity can be incorporated into the general framework of tensor decomposition, providing a better interpretation for physical meaning and facilitating many practical applications for tensor decomposition.

\subsection{Review of Hypergraph}

\begin{figure}[htb]  
      \centering  
      \includegraphics[width=0.9\linewidth]{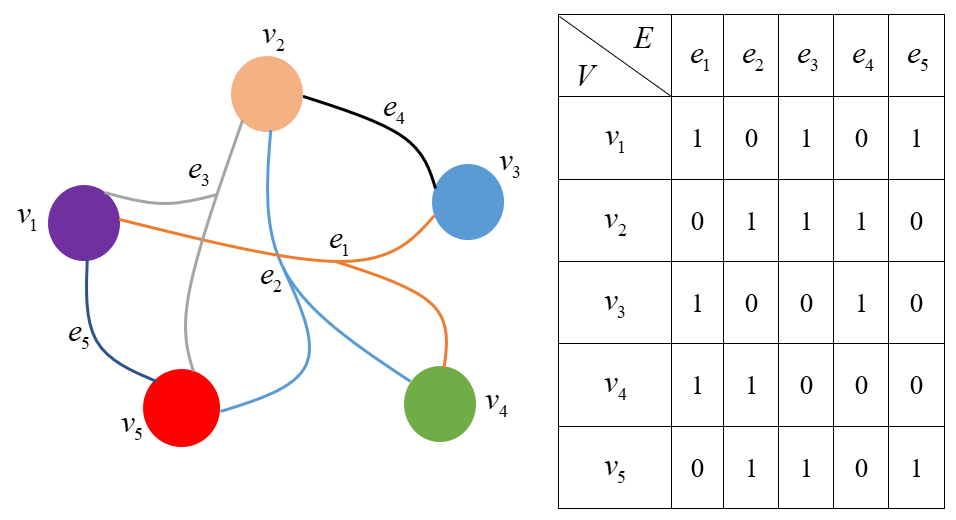} 
      \caption{Example of hypergraph and its incident relationship. In this hypergraph, an edge is a subset of vertexes. }
      \label{fig:hypergraph} 
\end{figure}

Here, we briefly review of hypergraph. Given a hypergraph $ G=\left ( V,E,W \right ) $, where $ V $ is the set of vertices, $ E $ is the set of hyperedges and $ W $ is weight matrix of hyperedge. Hence, each edge of hypergraph can be considered as a subset of vertices and connect more than two vertices (called hyperedge), whereas the edge of a 2-order graph only connects two vertices \cite{tian2009hypergraph}. We can use an indicator matrix $ H $ to express the relationship between vertices and hyperedges, that is defined by:

\begin{equation}\label{key}
h\left ( v,e \right )=\left\{\begin{matrix}
	1 & if\: \: \: v\in e\\ 
	0 & otherwise
\end{matrix}\right.
\end{equation}
\textcolor{black}{Note that $ v\in e $ and a vertex $ v\in V $, then a hyperedge $ e\in E $ is called an incident.} \textbf{Figure~\ref{fig:hypergraph}} is an example for hypergraph.

\section{HyperNTF} \label{hyperNTF}

\subsection{Modeling of HyperNTF}
\label{model}
\begin{figure}[htb]  
      \centering  
      \includegraphics[width=1.0\linewidth]{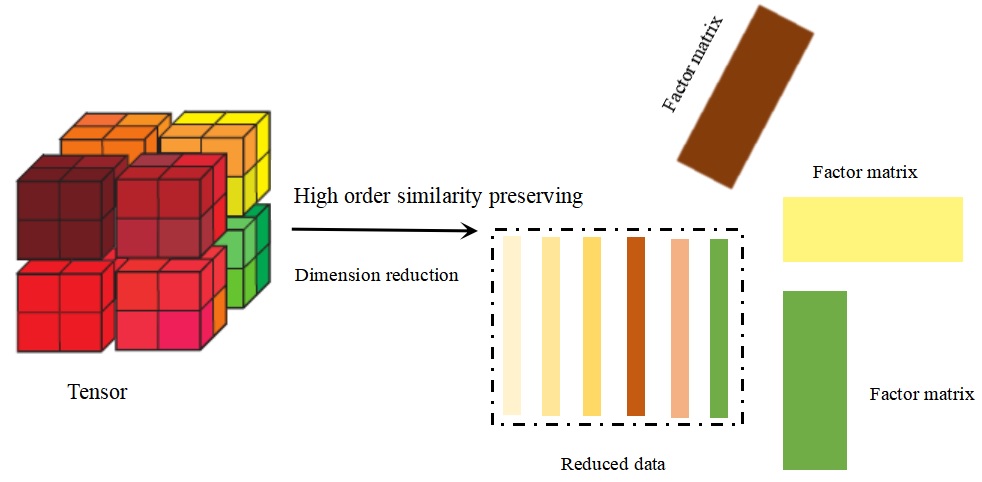} 
      \caption{HyperNTF decomposes input tensor into a set of factor matrices and reduced data, while maintaining the high-order similarity of nearest neighbors as much as possible.}
      \label{schematic} 
\end{figure}

As previous reviewed, non-negative tensor factorization is an effective tool for preserving non-negative property in dimensionality reduction, and hypergraph is an effective tool for modeling complex structures within original data. Therefore, to reveal the complex connections of nearest neighborhoods among samples, we incorporate hypergraph into the framework of non-negative tensor factorization (NTF). \textbf{Figure~\ref{schematic}} is the schematic illustration for Hypergraph Regularized Non-negative Tensor Factorization (HyperNTF). Note that the color denotes nearest similarity of geometric information.

Given non-negative objects $ \mathcal{X}\in \mathbb{R}^{L_1\times L_2\cdots\times M} $, $ \mathcal{X}\geqslant 0 $, to achieve the best reconstruction of tensor decomposition, we construct the following objective function:

\begin{equation}\label{eq:cost_HyperNTF}
\begin{aligned}
&\mathop{argmin}_{\left\{ U_n\right\}_{n=1}^{N-1}, Z} \underbrace{\left\|\mathcal{X}-\Sigma \times_1U_1\cdots\times _{N-1}U_{N-1}\times_NZ \right\|_F^2}\limits_{ \text{CP decomposition error term} }\\ 
& + \frac{\lambda }{2} \underbrace{\sum _{e\in E}\sum_{i,j\in V}\left \| Z_i - Z_j \right\|_F^2  \frac{w(e)h(i,e)h(j,e)}{d_E(e)}}\limits_{ \text{hypergraph regularization term} }\\
&s.t. \: \:\boldsymbol{1} U_n=\boldsymbol{1}, U_n\geq 0, n=1,\ldots, N-1; Z\geq 0\\
\end{aligned}
\end{equation}
where $ d_E\left ( e \right ) $ is the number of vertices (\textit{i.e.} $ i,j\in V $) incident with hyperedge (\textit{i.e.} $e$), that is calculated by $ d_E\left ( e \right )=\begin{matrix}\sum _{j\in V}\end{matrix}h\left ( j,e \right ) $. $\boldsymbol{1}$ is a row vector of all ones, and $ \Sigma\in \mathbb{R}^{J\times \cdots \times J} $ is a $ N \mathrm{th} $-order super-diagonal tensor. $ i $ and $ j $ are the index of samples. $ \lambda \geq 0 $ is the non-negative regularization parameter, controlling the importance of regularization term. 

In HyperNTF, the factor matrix (\textit{i.e.} $ Z $) associated with last mode of CP decomposition is utilized as the reduced data, in comparison to Tucker model that use the core tensor (\textit{i.e.} $ \mathcal{G} $) as the reduced representation, which can significantly reduce the storage consumption. Given a concrete example, in HyperNTF algorithm, it only needs to $ M\times J $ storage consumption for reduced data (\textit{i.e.} $ Z $). However, due to core tensor (\textit{i.e.} $ \mathcal{G} $) of Tucker decomposition is utilized as reduced representation, it requires to $ M{\textstyle \prod_{n=1}^{N-1}J_n} $ storage consumption for Tucker-based methods. 

\subsection{Solutions to HyperNTF}
To learn the factor matrices $ \left\{ U_n\right\}_{n=1}^{N-1} $ and its reduced data $ Z $ in Eq.~\eqref{eq:cost_HyperNTF}, we adopt Multiplicative Update Rule (MUR) to solve the learning objective ~\cite{lee2001algorithms,lee1999learning,li2016mr}. First, it needs to derive the gradient of Eq.~\eqref{eq:cost_HyperNTF} with respect to $ \left\{ U_n\right\}_{n=1}^{N-1} $ and $ Z $, thereby the first term of Eq.~\eqref{eq:cost_HyperNTF} is formulated as following:
\begin{equation}\label{eq:cost_HyperNTF_ref}
\begin{aligned}
&\mathop{argmin}_{U_1,\ldots, U_{N-1},Z}\left\|\mathcal{X}^{(n)}-U_n\left ( Z\odot _{i\neq n}U_i \right )^T \right\|_F^2\\
&=\left \| \mathcal{X}^{(n)} \right \|^2-2\left \langle \mathcal{X}^{(n)},U_n\left ( Z\odot_{i\neq n}U_i  \right )^T \right \rangle\\
&+\left \| U_n\left ( Z\odot_{i=N-1, i\neq n}^{i=1}U_i  \right )^T \right \|^2\\
\end{aligned}
\end{equation}
For simplify expression, we use $ Z\odot_{i\neq n}U_i=Z\odot U_{N-1}\odot \cdots \odot U_1 $ but except $ U_n $. Furthermore, the second term of Eq.~\eqref{eq:cost_HyperNTF} (i.e., hypergraph regularization) can be derived as following,

\begin{equation}\label{eq:term2}
\begin{aligned}
&\mathop{argmin}_{Z} \frac{1}{2}\sum _{e\in E}\sum_{i,j\in V}\frac{w(e)h(i,e)h(j,e)}{d_E(e)}\left \| Z_i-Z_j \right \|_F^2\\ 
=&\sum _{e\in E}\sum_{i,j\in V}\frac{w(e)h(i,e)h(j,e)}{d_E(e)}\left ( \left \| Z_i \right \|_F^2-\left \langle Z_i,Z_j \right \rangle \right )\\
=&\sum _{e\in E}\sum_{i\in V}w\left ( e \right )\left \| Z_i \right \|_F^2h\left ( i,e \right )\sum_{j\in V}\frac{h(j,e)}{d_E(e)}\\
&-\sum _{e\in E}\sum_{i,j\in V}\frac{w(e)h(i,e)h(j,e)}{d_E(e)}Z_iZ_j\\
=&\sum_{i\in V}\left \| Z_i \right \|_F^2d_V\left ( i \right )-Z^THWD_E^{-1}H^TZ\\
=&tr\left ( Z^T\left ( D_V-HWD_E^{-1}H^T \right )Z \right ) \\
\end{aligned}
\end{equation}
where $ D_V $ is the degree matrix of vertices, and $ D_E $ is the degree matrix of hyperedges, which are diagonal matrices; $ H $ is an indicator matrix, denoting relationship between vertex and hyperedge.

Since the objective function in Eq.~\eqref{eq:cost_HyperNTF} is a multi-variable, non-convex and constrained optimization problem, thus we incorporate Lagrange multipliers \textit{i.e.} $ \psi _n $, $ n=1,\ldots,N $ associated with each factor matrices, \textit{i.e.} $ U_n\geq 0 $, $ n=1,\ldots, N-1 $ and $ Z $ to relax the objective variables in Eq.~\eqref{eq:cost_HyperNTF}. As a consequence, the objective function in Eq.~\eqref{eq:cost_HyperNTF} can be formulated as following:

\begin{equation}\label{eq:Lagrange}
\begin{aligned}
&\mathcal{L} \left ( \left\{ U_n\right\}_{n=1}^{N-1},Z \right )\\
&=\left \| \mathcal{X}^{(n)} \right \|^2-2\left \langle \mathcal{X}^{(n)},U_n\left ( Z\odot_{i\neq n}U_i  \right )^T \right \rangle+\left \| U_n\left ( Z\odot_{i\neq n}U_i  \right )^T \right \|^2\\
& + \lambda tr\left ( Z^T\left ( D_V-HWD_E^{-1}H^T \right )Z \right )\\
& + \begin{matrix}\sum_{n=1}^{N-1}tr\left ( U_n^T\psi _n \right ) \end{matrix} + tr\left ( Z^T\psi _N \right )\\
\end{aligned}
\end{equation}

To this end, we can use an alternating procedure to update $ U_n $, $ n=1,\ldots,N-1 $ and $Z$. Furthermore, we derive the partial derivatives of Eq.~\eqref{eq:Lagrange} but except the first term of Eq.~\eqref{eq:Lagrange} since its partial gradient is a zero. Therefore, the second term of Eq.~\eqref{eq:Lagrange} is given by

\begin{equation}\label{key}
\left \langle \mathcal{X}^{(n)},U_n\left ( Z\odot _{i\neq n}U_i \right )^T \right \rangle=tr\left ( U_n\left ( Z\odot _{i\neq n}U_i \right )^T\mathcal{X}^{(n)T} \right )
\end{equation} 
And, the third term of Eq.~\eqref{eq:Lagrange} is

\begin{equation}\label{key}
\left \| U_n\left ( Z\odot _{i\neq n}U_i \right )^T \right \|_F^2=tr\left ( U_n\left ( Z^TZ\ast _{i\neq n}U_i^TU_i \right ) U_n^T\right ).
\end{equation} 

Recall that $ \left ( \odot _{i\neq n}U_i \right )^T\left ( \odot _{i\neq n}U_i \right )=\left ( \ast _{i\neq n}U_i^TU_i \right ) $, $ \frac{\partial }{\partial X}tr\left ( XU \right )=U^T $, and $ \frac{\partial }{\partial X}tr\left ( XUX^T \right )=XU^T+XU $, according to \cite{petersen2012matrix}. 

\textbf{Iterative Scheme with Factor Matrices}: To obtain the learning rule of $ U_n $ for $ n=1,\ldots, N-1 $, we first need to derive the partial gradient of Eq.~\eqref{eq:Lagrange} with $ U_n $, and obtain

\begin{equation}\label{eq:derivative_cost}
\frac{\partial f\left ( U_n \right )}{\partial U_n}=-2\mathcal{X}^{(n)}\left ( Z\odot _{i\neq n}U_i \right )+2U_n\left ( Z^TZ\ast _{i\neq n}U_i^TU_i \right )+\psi _n.
\end{equation} 
Then, by applying the Karush-Kuhn-Tucker (KKT) conditions (\textit{i.e.}, $ U_n\ast \psi _n=0 $ for $ n=1,\ldots N-1$), we can obtain solutions of Eq.~\eqref{eq:derivative_cost}, as following
\begin{equation}\label{eq:derivative_cost_KKT}
-2U_n\ast \left [ \mathcal{X}^{(n)}\left ( Z\odot _{i\neq n}U_i \right ) \right ]+2U_n\ast \left [ U_n\left ( Z^TZ\ast  _{i\neq n}U_i^TU_i \right ) \right ]=0
\end{equation} 
And, by using Eq.~\eqref{eq:derivative_cost_KKT}, we can obtain the solution with $ U_n $ from $ k\mathrm{th} $ iteration to the next $\left(k+1\right) \mathrm{th}$ iteration, as following

\begin{equation}\label{learning_rules1}
U_n^{k+1}\leftarrow \cdots U_n^k\ast \frac{\mathcal{X}^{(n)}\left ( Z\odot _{i\neq n}U_i \right )}{U_n\left ( Z^TZ\ast _{i\neq}U_i^TU_i \right )}
\end{equation}
Here, the Khatri-Rao product (\textit{i.e.} $ \odot $) of the involved $ U_i $ leads to a matrix of size $ \left ( \begin{matrix} \prod _{i\neq n}L_i\end{matrix}\times J \right ) $, which can get very costly in terms of computation and memory requirements when $ L_i $ and $ N $ are very large. To address such problem, we adopt an effective technique, called matricized tensor-times Khatri-Rao product (MTTKRP)~\cite{kaya2017high} to calculate the $ n\mathrm{th} $ mode vector multiplication, that is

\begin{equation}\label{learning_rules2}
A_j\leftarrow \mathcal{X}\times_1\mathrm{u}_{1,j}^T\cdots\times_{n-1}\mathrm{u}_{n-1,j}^T\times_{n+1}\mathrm{u}_{n+1,j}^T\cdots\times_N\mathrm{u}_{N,j}^T
\end{equation}

where $ \mathrm{u}_{n,j}^T $ is the column vector of factor matrix on the $ n\mathrm{th} $ mode, $ j $ is the index of columns, and $ T $ is the transpose of a matrix. As a result, the solution of $ A $ is calculated column by column can efficiently reduce the computational and storage consumption, resulting in the computation cost being the product of the tensor $ \mathcal{X} $ with $ \left(N-1 \right) $-vectors $ J $ times.

\begin{bfseries} Iterative Scheme with Reduced Data \end{bfseries}: By a similar way, we derive the learning rule of $ Z $, which involves with partial gradient of Eq.~\eqref{eq:Lagrange} with respect to $ Z $. In addition, and partial gradient of regularization term with respect to $ Z $. That is given by

\begin{equation}\label{learning_rules3}
\begin{aligned}
&\frac{\partial }{\partial Z}tr\left ( Z^T\left ( D_V-HWD_E^{-1}H^T \right )Z \right )\\
&=2\left ( D_V-HWD_E^{-1}H^T \right )Z\\
\end{aligned}
\end{equation}
Then, by applying the KKT conditions (i.e., $ Z\ast \psi _N=0 $), we can obtain the update strategy concerning $ Z $ expressed by the form of tensor mode multiplied by a set of vectors. Or equivalently, it can be expressed as the following iterative scheme,

\begin{equation}\label{learning_rules4}
Z^{k+1}\leftarrow \cdots Z^k\ast \frac{\mathcal{X}^{(n)}\left ( \odot _{i\neq N}U_i \right )+HWD_E^{-1}H^TZ}{Z\left ( \ast _{i\neq N}U_i^TU_i \right )+D_VZ}
\end{equation}

Hence, the learning rule of $U_n$ and $Z$ are presented in Eq.~\eqref{learning_rules1} and Eq.~\eqref{learning_rules4} respectively. We use the Multiplicative Update Rule (MUR) to solve this objective function. Specifically, we first randomly initiate the factor matrices $ \left\{ U_n\right\}_{n=1}^{N-1} $ and $ Z $, and then iterative updating them by Eq.~\eqref{learning_rules1} and Eq.~\eqref{learning_rules4}, until the termination criteria are met. At each iteration, we update matrix for one mode with fixing matrix for other modes, resulting in a new objective function that depends only on the specific mode to be learned. After learning all the matrices $ \left\{ U_n\right\}_{n=1}^{N-1} $ and $ Z $, we record the total number of iterations and examine the convergence at the end of each iteration. The pseudo-code of HyperNTF is given in \textbf{Algorithm~\ref{alg1}}, and the Matlab code of HyperNTF is available at https://github.com/ncclabsustech/HyperNTF.

\begin{algorithm} 
	\caption{Hypergraph Regularized Nonnegative Tensor Factorization (HyperNTF)} 
	\label{alg1} 
	\begin{algorithmic}[1] 
		\REQUIRE Input data $ \mathcal{X}\in \mathbb{R}_+^{L_1\times\cdots\times L_{N-1}\times M} $, regularization parameter $ \lambda $, and maximum iteration $ \mathrm{\bf{maxiter}} $\\
		\STATE $ \left\{ U_n\right\}_{n=1}^{N-1} $ and $ Z $, initial factor matrices
		\STATE repeat $ \mathrm{\bf{maxiter}} $
		\STATE update the factor matrices $ U_n $ by Eq. (\ref{learning_rules1})
		\STATE normalize columns of $ U_n $
		\STATE update the reduced data $ Z $ by Eq. (\ref{learning_rules4}).
		\STATE until termination criteria $ RSE_k< Tol $ or the maximum number of iterations is met.
		\STATE OUTPUT $ Z\in \mathbb{R}_+^{M\times J} $
	\end{algorithmic} 
\end{algorithm}

\subsection{Computational Complexity}

In this subsection, we analyze the computational complexity of HyperNTF. First, for the construction of hypergraph, it needs to construct the $ k $-nearest neighboring graph of weight matrix $ W $. The computational complexity of generating $ W $ needs to perform operations of $ MK\prod_{n=1}^{N-1} L_n $ multiplication and addition. Recall that the weight matrix is a sparsity matrix, the average nonzero elements on each row of $ W $ is $ K $, and $ K $ is the selection of nearest neighbors. In addition, the hypergraph regularization term involves with calculation of $ HWD_E^{-1}H^TZ $ and $ D_VZ $, thereby it needs to perform $ M^4 $ and $ M $ arithmetic operations respectively.

Then, we analyze the computational complexity for Eq. (\ref{learning_rules1}) and Eq. (\ref{learning_rules4}), including the matricized tensor-times Khatri-Rao product (MTTKRP)~\cite{kaya2017high}, matrix product, Hadamard product, and tensor mode unfolding.

\begin{bfseries} Iterative Scheme with Factor Matrices \end{bfseries}: For the computation of MTTKRP \textit{i.e.} $ \mathcal{X}^{(n)}\left ( Z\odot _{i\neq n}U_i \right ) $, it yields $ A\in \mathbb{R}^{L_n\times J} $ with column vector, as following

\begin{equation}\label{MTTKRP_1}
A_j = \mathcal{X}\times_1\mathrm{u}_{1,j}^T\cdots\times_{n-1}\mathrm{u}_{n-1,j}^T\times_{n+1}\mathrm{u}_{n+1,j}^T\cdots\times_N\mathrm{z}_j^T
\end{equation}
It has a total of $ J $ column vectors, and the generation of each column vector needs to conduct $ M\prod_{i=1,i\neq n}^{N-1}L_i $ multiplication, thereby for each mode calculation of MTTKRP needs to perform $ MJ\prod_{i=1,i\neq n}^{N-1}L_i $ arithmetic operations. Since the multiplication plays the most important role in computational complexity, in the following, we focus on the multiplication operations. The calculation of $ U_n\left ( Z^TZ\ast _{i\neq n}U_i^TU_i \right ) $ needs to perform $ J^{2(N-1)} $ arithmetic operations. Hence, the computational complexity of Eq. (\ref{learning_rules1}) is given by

\begin{equation}\label{MTTKRP_1}
O_1 = MJ\begin{matrix}\prod_{i=1,i\neq n}^{N-1}L_i\end{matrix} + J^{2(N-1)}
\end{equation}
We can see that the computational complexity of Eq. (\ref{learning_rules1}) is linear with number of training samples, and exponentially with input dimension and reduced dimension.

\begin{bfseries} Iterative Scheme with Reduced Data \end{bfseries}: By a similar way, we analyze the computational complexity of Eq. (\ref{learning_rules4}). For the calculation of MTTKRP \textit{i.e.} $ \mathcal{X}^{(n)}\left ( \odot _{i\neq n}U_i \right ) $, it yields $ A\in \mathbb{R}^{L_n\times J} $ with column vector, as following

\begin{equation}\label{MTTKRP_2}
A_j = \mathcal{X}\times_1\mathrm{u}_{1,j}^T\cdots\times_{N-1}\mathrm{u}_{N-1,j}^T
\end{equation}
It has a total of $ J $ column vectors, and the generation of each column vector needs to conduct $ \prod_{i=1}^{N-1}L_i $ multiplication, thereby for each mode calculation of MTTKRP needs to perform $ J\prod_{i=1}^{N-1}L_i $ arithmetic operations. Additionally, the calculation of $ Z\left ( \ast _{i\neq N}U_i^TU_i \right ) $ needs to perform $ J^{2(N-1)} $ arithmetic operations.

\begin{equation}\label{key}
M\begin{matrix}\sum_{m=1,m\neq n}^{N-1}\end{matrix}\left ( \begin{matrix}\prod_{j=1,j\neq n}^m J_j\end{matrix}\right )\left ( \begin{matrix}\prod_{i=m,i\neq n}^{N-1} L_iL_n\end{matrix}\right )
\end{equation}
Therefore, the computational complexity of Eq. (\ref{learning_rules4}) is given by

\begin{equation}\label{MTTKRP_1}
O_2 = MK\begin{matrix}\prod_{n=1}^{N-1} L_n\end{matrix} + M^4 + M + J\begin{matrix}\prod_{i=1}^{N-1}L_i\end{matrix} + J^{2(N-1)}
\end{equation}
which is exponentially with number of training samples, input dimension, and reduced dimension.

\section{Experiments}
To validate the effectiveness of hypergraph and HyperNTF for dimensionality reduction, we conduct two types of experiments, including the unfolding of synthetic manifold, clustering images, and classification EEG. All experiments were conducted on a desktop computer with an Intel Core i5-5200U CPU at 2.20GHz and 8.00 GB RAM, and repeated 10 times to reduce variability, with objects randomly selected each time.

\subsection{Manifold unfolding}
\label{manifold unfolding}
In the unfolding of synthetic manifold experiments, the hypergraph is utilized to encode the complex connections among samples, then the unfolding operation follows the general procedures of manifold learning as below ~\cite{belkin2003laplacian, zhang2004principal}: Firstly, we calculate the local similarity connections by using $ k $-nearest neighboring search, and calculate the degree matrix of vertices and hyperedges; Secondly, we extract the principal eigenvectors by hypergraph Laplacian; Finally, visualizing the geometric information of nearest neighborhoods in 2-D space.
\subsubsection{Simulated 3D manifolds}
We conduct the manifold unfolding test by using the simulated data Matlab Demo (mani.m)~\cite{qiao2012explicit}. Firstly, the simulated manifolds (\textit{i.e.} Punctured Sphere, Gaussian surface, Twin Peak, and Toroidal Helix) is generated in the three-dimensional space. On each manifold, 1000 data samples are randomly generated for training, we set the nearest neighbor $ k=44 $ for Punctured Sphere, $ k=25 $ for Gaussian surface, $ k=15 $ for Twin Peaks, and $ k=10 $ for Toroidal Helix respectively, the polynomial degree is set to $ p=2 $. 

\subsubsection{Hypergraph for manifold unfolding}
To measure the high-order similarity among samples, we incorporate hypergraph into the procedures of manifold learning, and then extract the principal eigenvectors to project these 3D manifolds to the 2D space. We show the unfolding results for Punctured Sphere, Gaussian surface, Twin Peaks, and Toroidal Helix in \textbf{Figure 3-6}, respectively.

\begin{figure}[htb]  
      \centering  
      \includegraphics[width=0.9\linewidth]{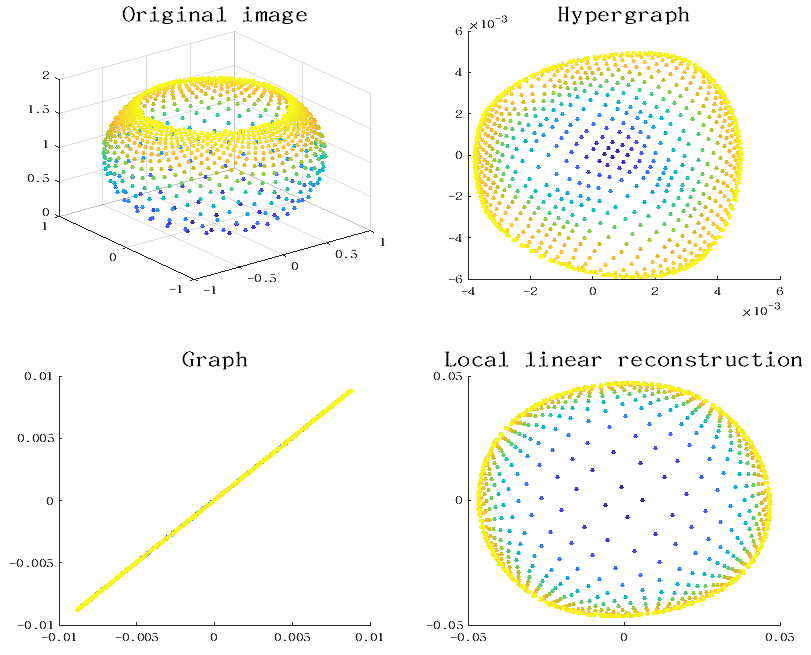} 
      \caption{The experimental results of unfolding Punctured Sphere embedded in $ \mathbb{R}^3 $, using Hypergraph, Graph and local linear reconstruction by LLE. The learning result of Hypergraph is better than Graph and local linear reconstruction by LLE.}
      \label{fig:Sphere} 
\end{figure}

\begin{figure}[htb]  
      \centering  
      \includegraphics[width=0.9\linewidth]{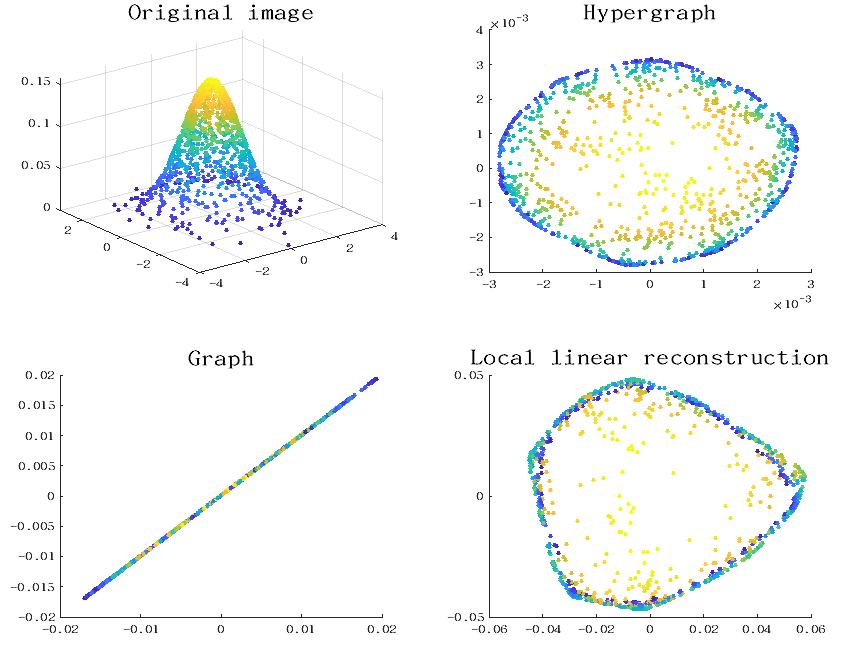} 
      \caption{\textcolor{black}{Experiments on unfolding Gaussian surface embedded in $ \mathbb{R}^3 $, learning result of Hypergraph is better than Graph and local linear reconstruction by LLE.}}
      \label{fig:Gaussian} 
\end{figure}

As shown in \textbf{Figure~\ref{fig:Sphere}}, the higher order similarity among samples measured by hypergraph can effectively preserve the topological structures in dimensionality reduction for Punctured Sphere. In comparison to hypergraph, using local linear reconstruction by LLE can only preserves well on the boundary position, but it is sparsity on the centring position of Punctured Sphere, which demonstrate that a lot of significant information have been lost in dimensionality reduction. Unfortunately, the topological structure measured by graph is disrupt largely. All of these indicate that for unfolding the Punctured Sphere, hypergraph is superior to the graph and local linear reconstruction by LLE algorithm. As shown in \textbf{Figure~\ref{fig:Gaussian}}, for unfolding of the Gaussian surface show a similar result, which demonstrate that hypergraph is robust in many practical applications.

\begin{figure}[htb]  
      \centering  
      \includegraphics[width=0.9\linewidth]{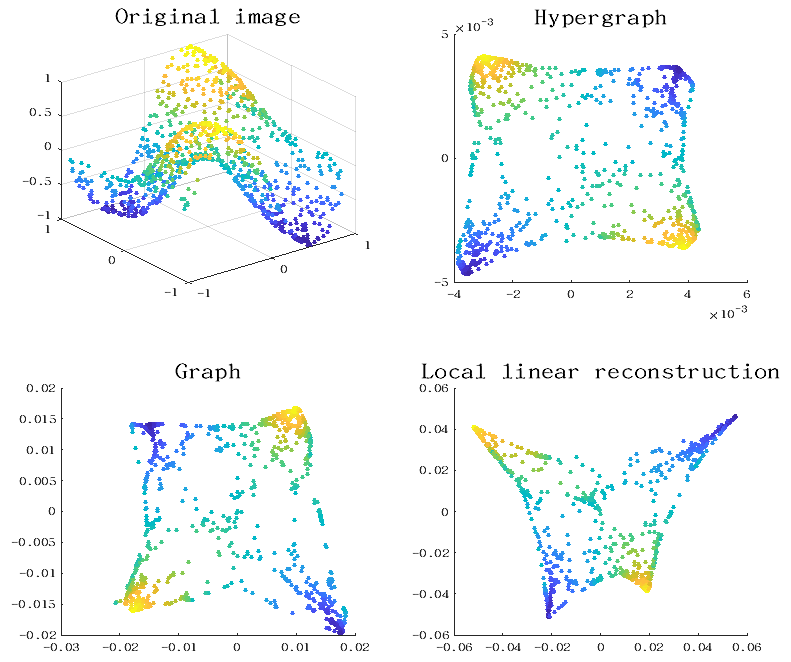} 
      \caption{Experiments on unfolding Twin Peaks embedded in $ \mathbb{R}^3 $, learning result of Hypergraph is better than Graph and local linear reconstruction by LLE.}
      \label{fig:Peaks} 
\end{figure}

All three methods in \textbf{Figure~\ref{fig:Peaks}} can effectively maintain the symmetric structure of original image, and hypergraph gets best performance for preserving the topological information in dimensionality reduction. However, for unfolding the Twin Peaks, using the local linear reconstruction will lost a lot of significant information.

\begin{figure}[htb]  
      \centering  
      \includegraphics[width=0.9\linewidth]{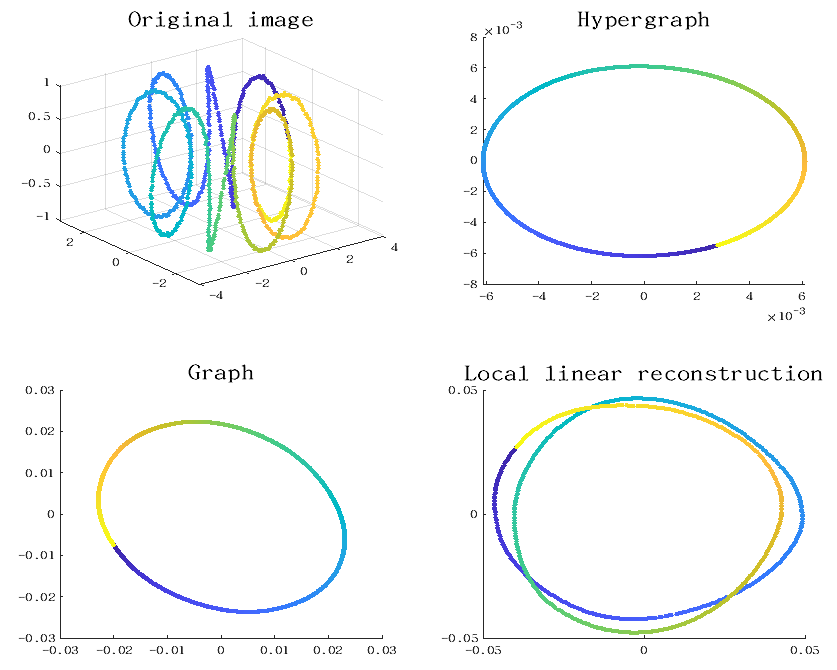} 
      \caption{Experiments on unfolding Toroidal Helix embedded in $ \mathbb{R}^3 $, learning result of Hypergraph is better than Graph and local linear reconstruction by LLE.}
      \label{fig:Helix} 
\end{figure}

From \textbf{Figure~\ref{fig:Helix}}, we can see that hypergraph and graph have a closely similarity performance for dimension reduction of Toroidal Helix, while the contour obtained by local linear reconstruction is largely distorted.

To summarize, measuring the complex structures by hypergraph can effectively preserve the topological information in dimensionality reduction, if only an appropriate $ k $-nearest neighbors is selected. All results consistently show that hypergraph is superior to graph and local linear reconstruction by LLE algorithm, for revealing the higher-order similarity among data points and recovering the underlying structures of two degrees of freedom.

\subsection{Image Clustering} \label{cluster}

\subsubsection{Illustration of image datasets}
In the first cluster experiment, five image datasets (\textit{i.e.} COIL20, ETH80, MNIST Digits, Olivetti Faces, and USPS) are utilized. The data are randomly shuffled, and the gray value of pixels is normalized to the unit. Each dataset used in our cluster analysis has the ground-truth class label. \textcolor{black}{For evaluation, we first reduce the dimension of dataset and then cluster them with $ k $-means algorithm.} Therefore, 3-order tensors are used to execute our cluster experiments, the first two modes are associated with image pixels, and the last mode is associated with number of image data.

The used COIL20 dataset contains 1420 grayscale images of 20 objects viewed from 72 equally spaced orientations. The images contain $ 32\times 32=1024 $ pixels. The ETH80 is a multi-view image dataset for object categorization, which includes eight categories that include eight categories correspond to apple, car, cow, cup, dog, horse, pear and tomato. Each category contains ten objects, and each object is represented 41 images of different views. The original image resolution is $ 128\times 128 $, each image was to be resized $ 32\times 32 $ pixels, for a total of 3280 images. The MNIST dataset contains 60000 grayscale images of handwritten digits. For our experiments, we randomly selected 3000 of the images for computational reasons. The digit images have $ 28\times 28=784 $ pixels. The Olivetti faces dataset consists of images of 40 individuals with small variations in viewpoint, large variations in expression, and occasional glasses. The dataset consists of 400 images (10 per individual) of size $ 64\times 64=4096 $ pixels and is labeled according to identity. The USPS is a handwritten digits dataset, which contains a total of 2000 images of size $ 16\times 16=256 $ pixels.

\textbf{Table 2} presents the general description of the datasets used in cluster analysis, wherein $Size_{original}$ refers to the raw image size, $Size_{reduced}$ refers to the size of the dataset after dimensionality reduction.


\begin{table}
    \caption{Illustrations of the datasets}
    \label{table:table2}
    \centering
    \small
    \begin{center}
    \scalebox{0.90}{
    \begin{tabular}{lcccc}
    \toprule
        dataset & \#samples & size$_{original}$ & size$_{reduced}$ & \#classes\\
    \midrule
        COIL20 & 1440 & 32*32 & 1440*32 & 20 \\
    	ETH80 & 328 & 32*32 & 328*32 & 8 \\ 
    	MNIST & 3000 & 28*28 & 3000*28 & 10 \\ 
    	USPS & 2000 & 16*16 & 2000*16 & 10 \\ 
    	Olivetti & 400 & 64*64 & 400*32 & 40 \\ 
    \bottomrule
\end{tabular}}
\end{center}
\end{table}

\subsubsection{Parameters selection}

\begin{figure}[htb]  
      \centering  
      \includegraphics[width=1\linewidth]{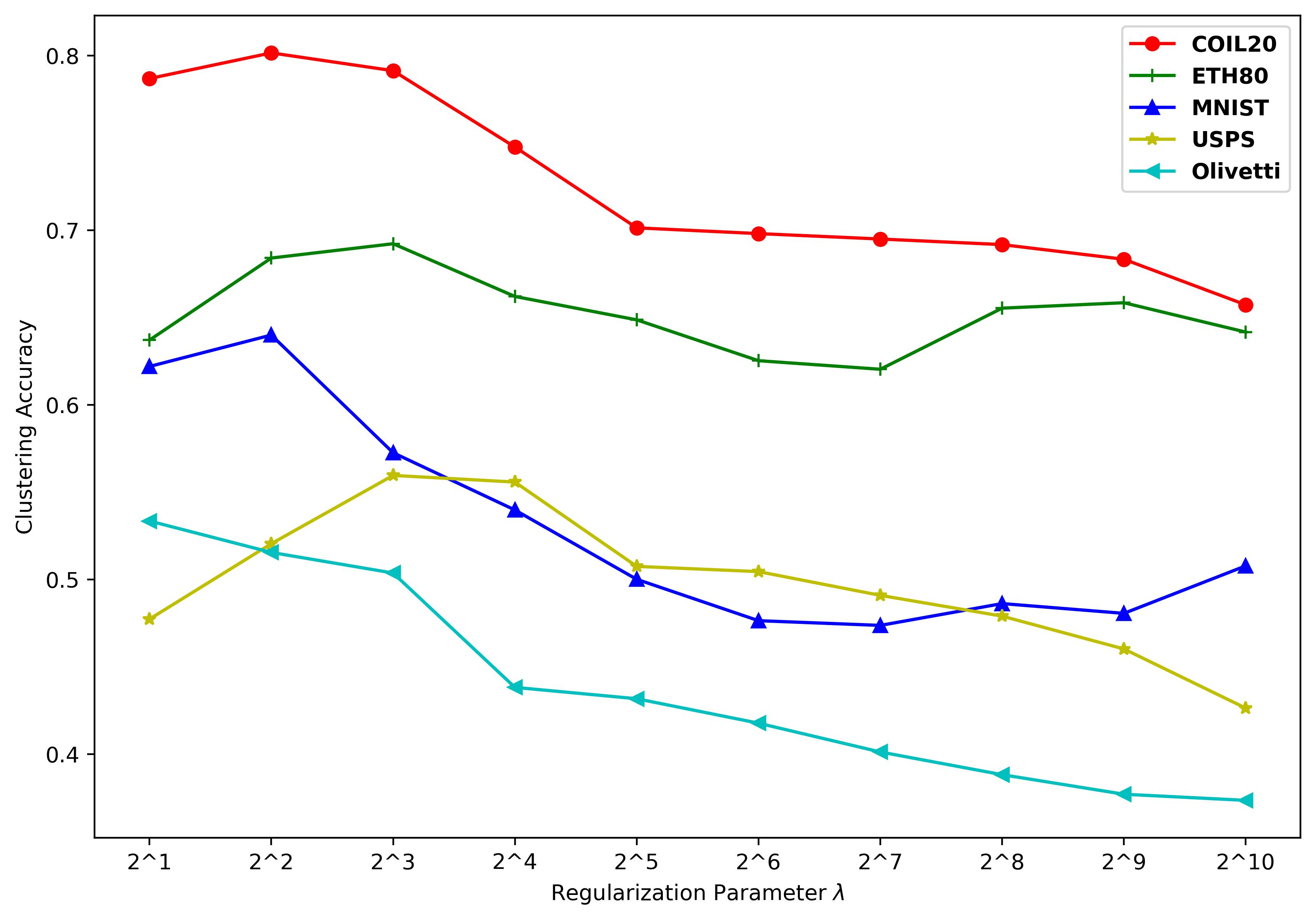}
      \caption{The parameter selection of cluster analysis for $ \lambda $ varies from $ 2^1 $ to $ 2^{10} $, and simultaneously varies $ k $ from $ 2\times 1 $ to $ 2\times 10 $. The selected parameters for the subsequent cluster analysis are indicated by the round markers.}
      \label{fig:parameter} 
\end{figure}

Here, we present the experimental results on the cluster analysis. Since HyperNTF involves two essential parameters, the regularization parameter $ \lambda $ and $ k $-nearest neighbors. To test the effects of regularization parameters selection $ \lambda $, and $ k $-nearest neighbors for constructing the connections between samples, \textcolor{black}{we vary $ \lambda $ from $ 2^1 $ to $ 2^{10} $ and simultaneously vary $ k $ in range from $ 2\times 1 $ to $ 2\times 10 $.} To this end, we run $ k $-means clustering 10 times with random initialization $ U_n $, $ n=1,\ldots, N-1 $ and $ Z $, then compute the averaged results as the final clustering results. We use the clustering Accuracy (Acc), and Normalized Mutual Information (NMI) as two evaluation metrics. As shown in \textbf{Figure \ref{fig:parameter}}, the performance of HyperNTF is robust with different selection of $ \lambda $ and $ k $-nearest neighbors.

Following to \textbf{Figure \ref{fig:parameter}}, we set parameters $ \lambda=4 $ \& $ k=3 $ for COIL20, $ \lambda=2 $ \& $ k=2 $ for ETH80, $ \lambda=2 $ \& $ k=3 $ for MNIST, $ \lambda=3 $ \& $ k=3 $ for Olivetti, and $ \lambda=4 $ \& $ k=3 $ for USPS respectively.

\subsubsection{Convergence}
\begin{figure}[htb]  
      \centering  
      \includegraphics[width=0.92\linewidth]{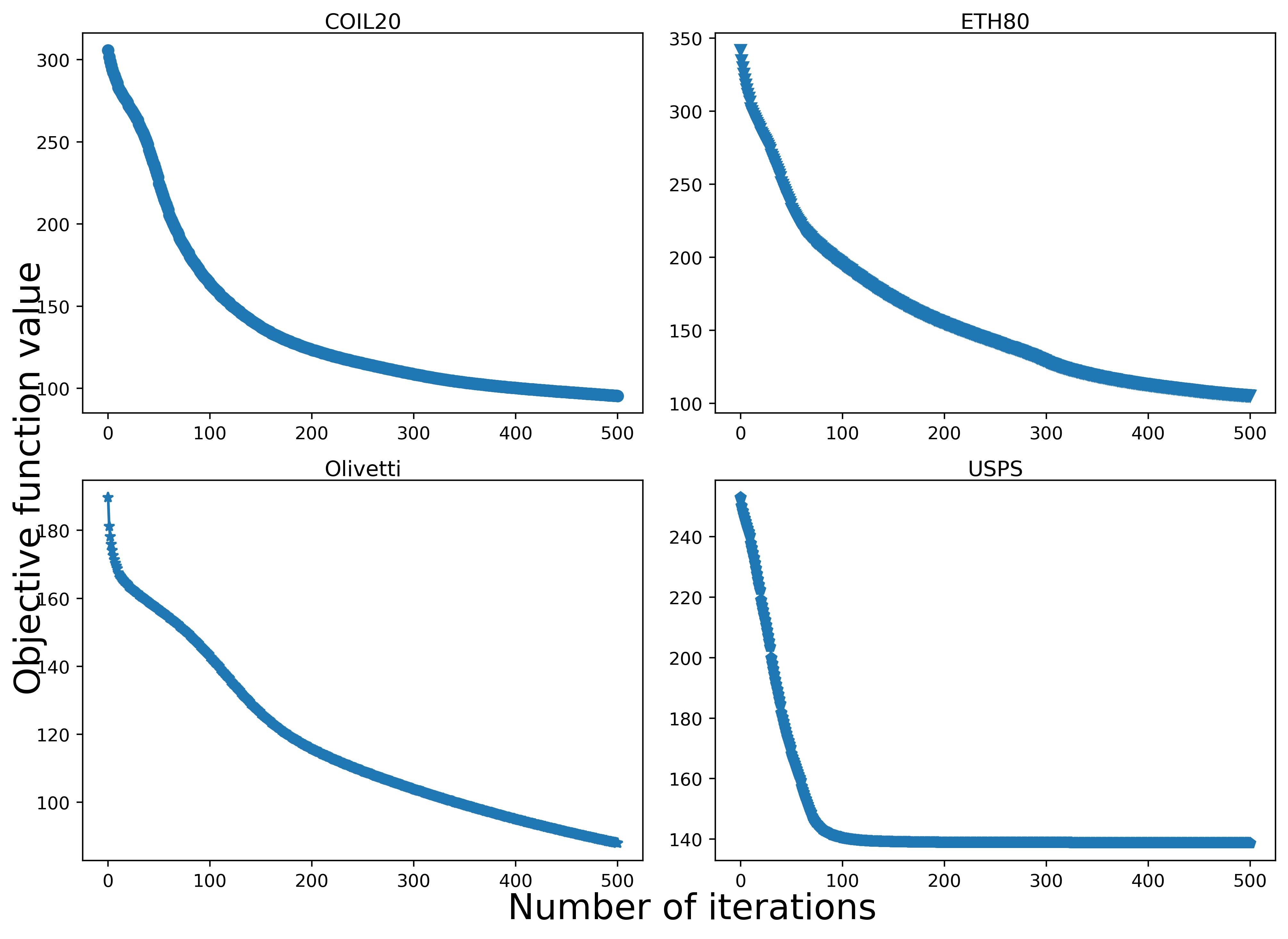} 
      \caption{\textcolor{black}{Demonstration of convergence of the proposed HyperNTF algorithm on different datasets.}}
      \label{Convergence} 
\end{figure}

\textcolor{black}{
Here, we investigate and demonstrate the convergence of HyperNTF by using the iteratively updating rules of factor matrices in Eq.~\eqref{learning_rules1} and its reduced data in Eq.~\eqref{learning_rules4}. Our stopping criterion is $ \left | \mathcal{O}_{k+1} - \mathcal{O}_k\right | < 0.1 $. We show the convergence curves of the HyperNTF on four image datasets in \textbf{Figure~\ref{Convergence}}. As shown in \textbf{Figure~\ref{Convergence}}, with increasing of iteration number, the objective function value is decreased.}

\subsubsection{Comparison with state-of-the-art methods}
To validate the effectiveness of our proposed HyperNTF, we compare HyperNTF with six existing methods, including the higher order singular value decomposition (HOSVD) \cite{de2000multilinear,savas2007handwritten}, nonnegative Tucker decomposition (NTD) \cite{kim2007nonnegative}, non-negative tensor factorization (NTF) \cite{cichocki2007nonnegative}, heterogenous tensor decomposition (HTD-Multinomial) ~\cite{sun2015heterogeneous}, low-rank regularized heterogeneous tensor decomposition (LRRHTD) \cite{zhang2017low}, and \textcolor{black}{graph-Laplacian Tucker decomposition (GLTD) \cite{jiang2018image}}.

\begin{figure}[htb]  
      \centering  
      \includegraphics[width=0.92\linewidth]{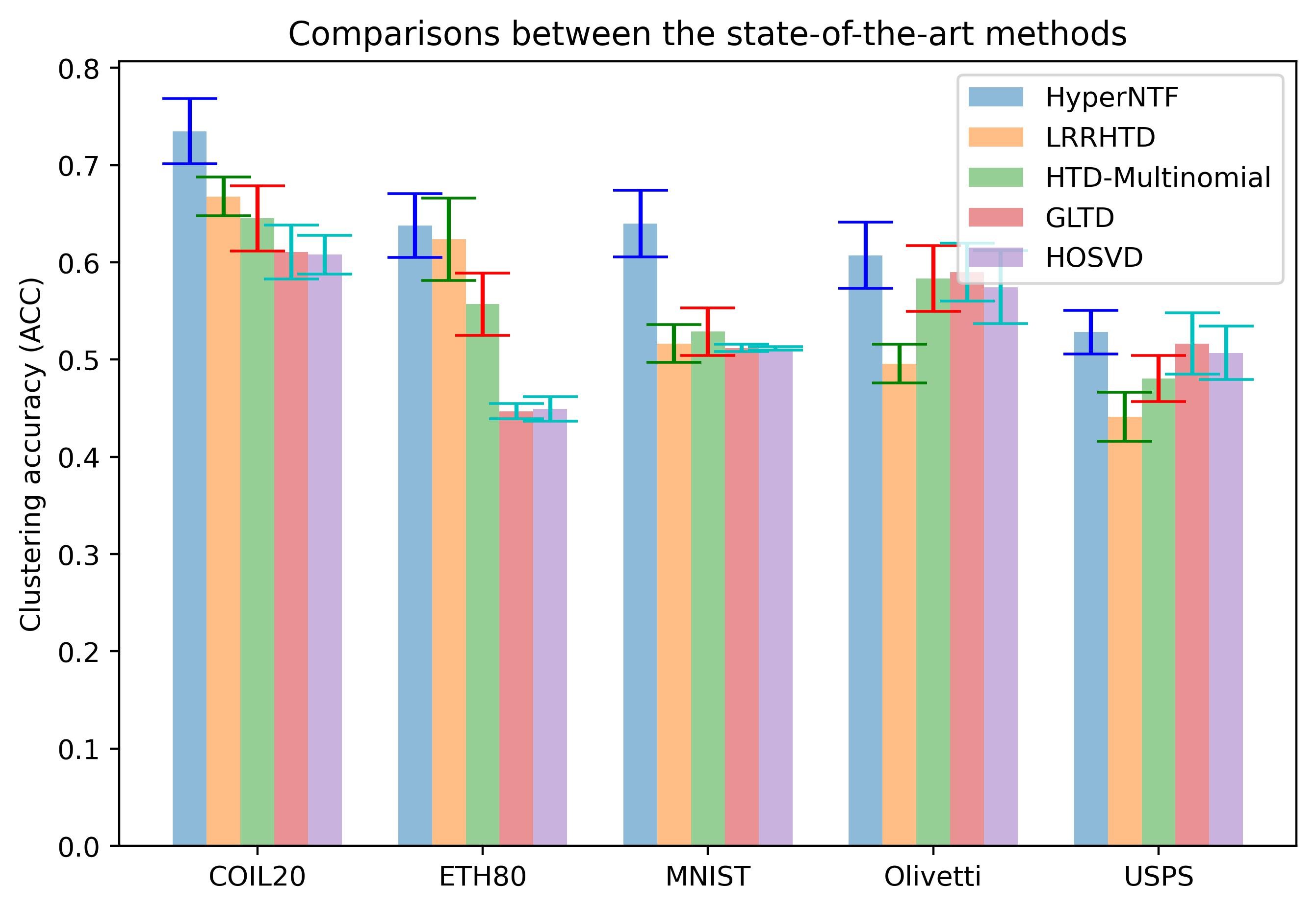} 
      \caption{The average and standard deviation of clustering accuracy (ACC) using $k$-means cluster by HyperNTF, LRRHTD, HTD-Multinomial, GLTD and HOSVD algorithms on the five image datasets.}
      \label{clustering_ACC} 
\end{figure}
As shown in \textbf{Figure~\ref{clustering_ACC}}, the clustering accuracy of HyperNTF is higher than the comparison algorithms, including LRRHTD, HTD-Multinomial, GLTD, and HOSVD, which demonstrate that HyperNTF can preserve the significant information in dimensionality reduction.

\begin{figure}[htb]  
      \centering  
      \includegraphics[width=0.92\linewidth]{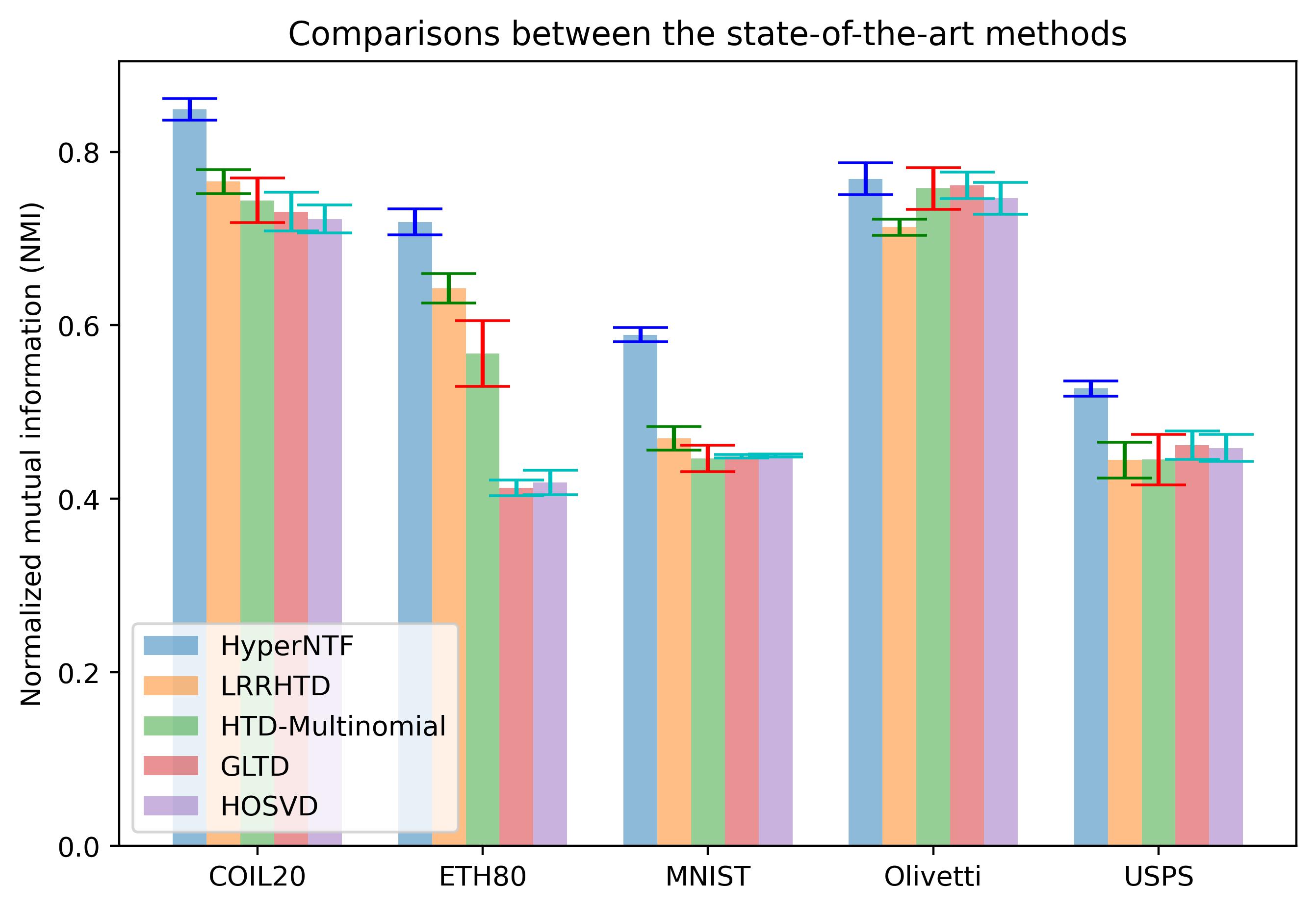} 
      \caption{\textcolor{black}{The mean and standard deviation of normalized mutual information (NMI) using $ k $-means cluster by HyperNTF, LRRHTD, HTD-Multinomial, GLTD and HOSVD algorithms on the five image datasets.}}
      \label{clustering_NMI} 
\end{figure}
From \textbf{Figure~\ref{clustering_NMI}}, we can see that the normalized mutual information (NMI) of HyperNTF is better than the comparison algorithms as well. By using two metrics, we show that HyperNTF is very suitable for clustering of image data.

\subsubsection{Regularization effects}

\begin{figure}[htb]  
      \centering  
      \includegraphics[width=0.92\linewidth]{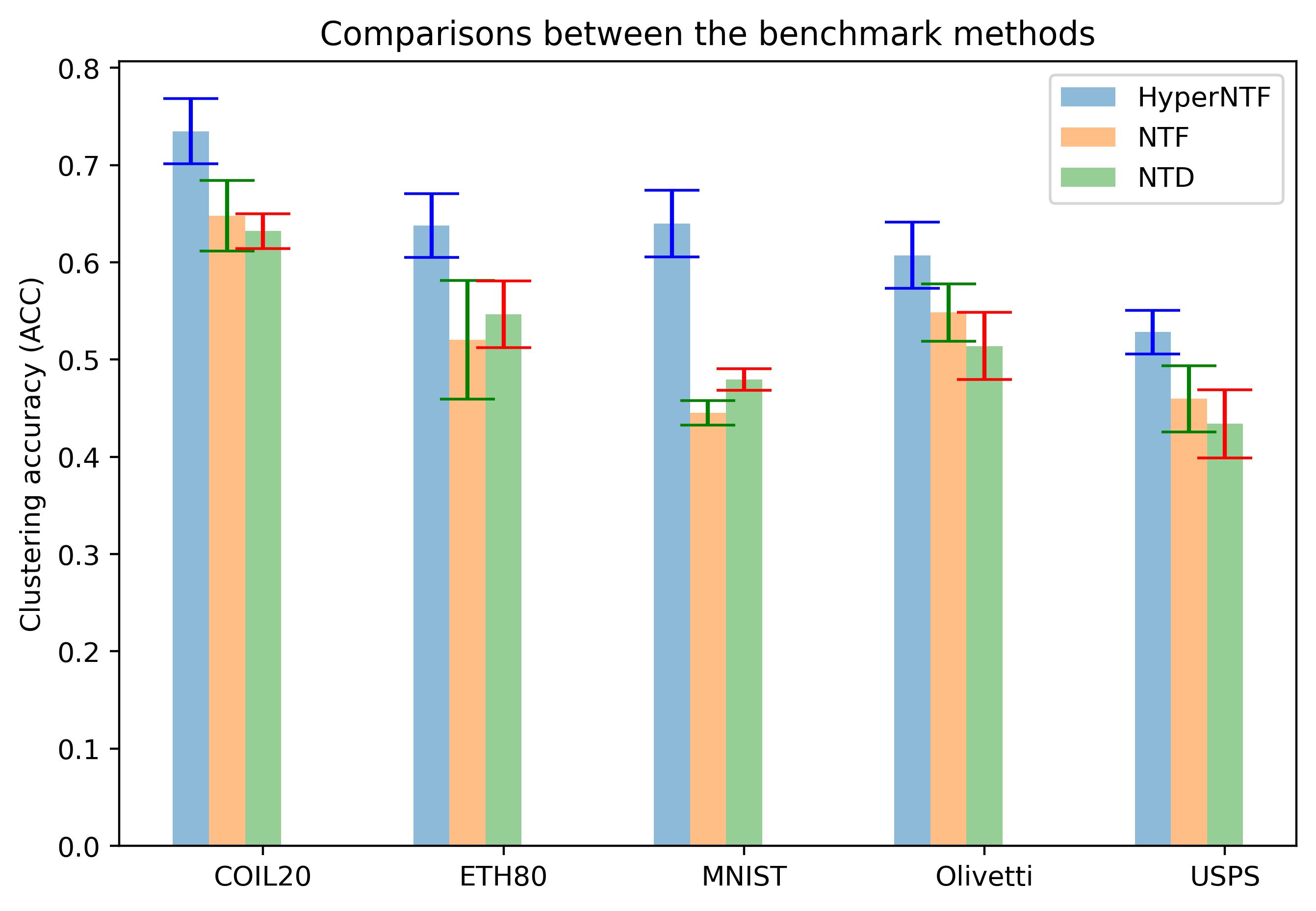}
      \caption{The average and standard deviation of clustering accuracy (ACC) using $ k $-means cluster by HyperNTF, NTF and NTD algorithms on the five image datasets.}
      \label{ACC_ablation} 
\end{figure}

\begin{figure}[htb]  
      \centering  
      \includegraphics[width=0.92\linewidth]{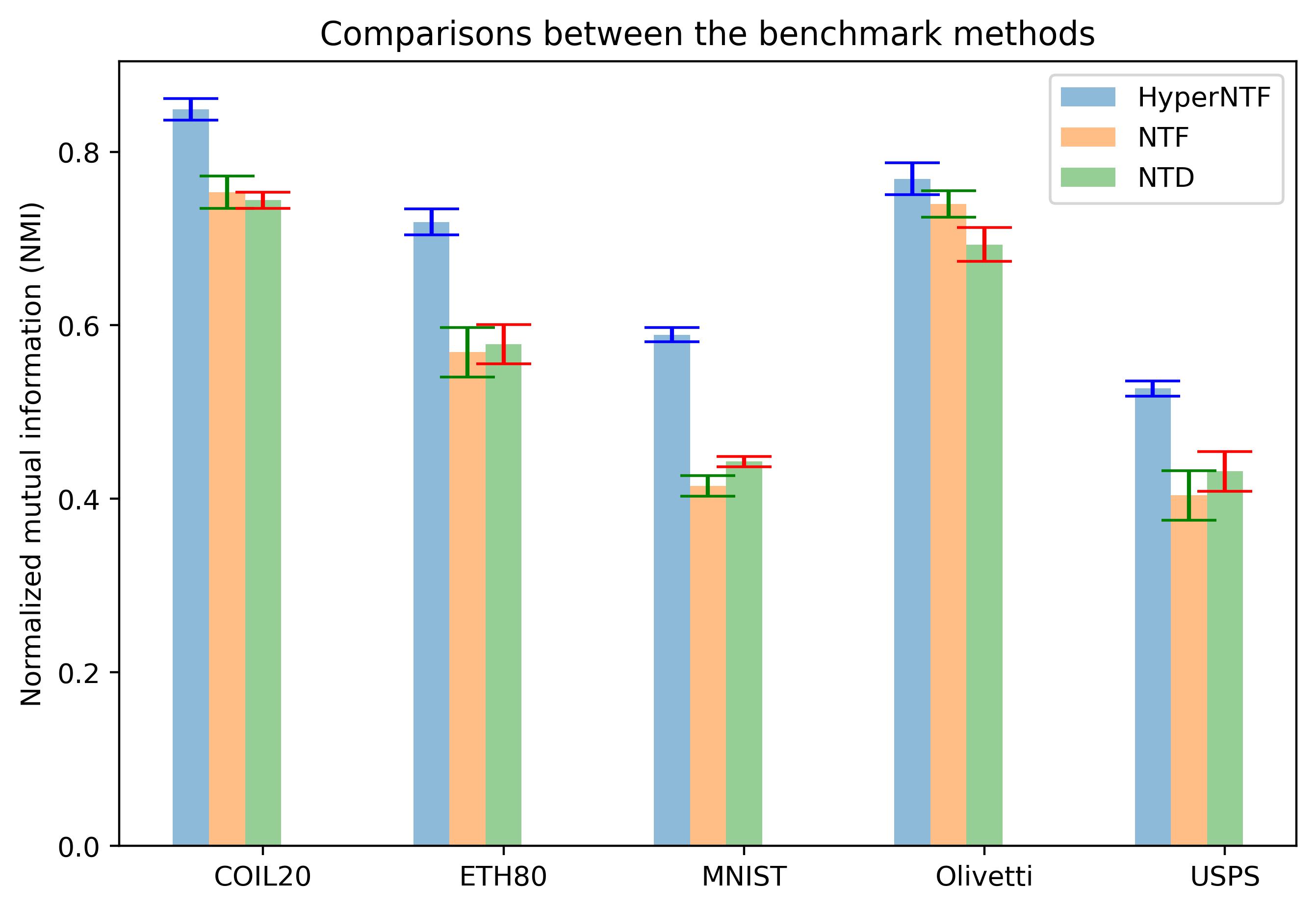} 
      \caption{The average and standard deviation of normalized mutual information (NMI) using $ k $-means cluster by HyperNTF, NTF and NTD algorithms on the five image datasets.}
      \label{NMI_ablation} 
\end{figure}

To validate the effectiveness of the hypergraph regularization term in Eq.~\eqref{eq:cost_HyperNTF}, we compare HyperNTF with standard methods, \textit{i.e.} non-negative tensor factorization (NTF) ~\cite{cichocki2007nonnegative} and non-negative Tucker decomposition (NTD)~\cite{kim2007nonnegative}, whereas NTF is the regularization parameter of HyperNTF reduced to zero. We use two metrics (\textit{i.e.} ACC and NMI) to evaluate their clustering results. \textbf{Figure~\ref{ACC_ablation}} and \textbf{Figure~\ref{NMI_ablation}} show their clustering results.

As shown in \textbf{Figure~\ref{ACC_ablation}}, HyperNTF can reliably cluster the data into the labeled classes regardless of the different number of cluster labels.

Analogously, as shown in \textbf{Figure~\ref{NMI_ablation}}, by incorporating the hypergraph into the general framework of non-negative tensor decomposition (\textit{i.e.} NTD and NTF) provides an improvement on the performance evaluation by using normalized mutual information (NMI).

\subsection{EEG signal analysis}\label{classification}
\subsubsection{EEG data and pre-processing} Electroencephalography (EEG) records brain activities as multichannel time series from multiple electrodes placed on the scalp of a subject to provide a direct communication channel between brain and computer, which is widely used in noninvasive brain-computer interfaces (BCI) applications. Here, we use the public PhysioNet motor imagery (MI) dataset and Macau steady-state visual evoked potential (SSVEP) dataset to perform classification experiments. 

The used PhysioNet EEG MI dataset consits of 2-class MI tasks (\textit{i.e.} runs 3, 4, 7, 8, 11, and 12, with imagine movements of left fist or right fist) ~\cite{schalk2004bci2000}, which is recorded from 109 subjects with 64-channel EEG signals (sampling rate equals to 160 Hz) during MI tasks. We randomly select 10 subjects from PhysioNet MI dataset in our experiments. The EEG signals are filtered with a band-pass filter (cutoff frequencies at $ 7\sim35 $ Hz) and a spatial filter (\textit{i.e.} Xdawn with 16 filters), therefore the resulting data is represented by $ trials \times channel \times time $. 

The Macau steady-state visual evoked potential (SSVEP) signals were recorded by our collaborators at University of Macau with ethical approval, containing 128-channel EEG recordings from 7 subjects sampled at 1000 Hz. The SSVEP dataset contains about two types of visual stimuli. The raw EEG data were pre-processed by sampling to 200Hz and then segmented into epochs (1-1000ms for SSVEP datasets). We use the wavelet transform to convert EEG signals into 4-order tensor representations in the trail-spatial-spectral-temporal domain (\textit{i.e.} $ 390\times 20 \times 20 \times 20 $ for SSVEP data, respectively).

\subsubsection{EEG classification results}

\begin{figure}[htb]  
      \centering  
      \includegraphics[width=0.95\linewidth]{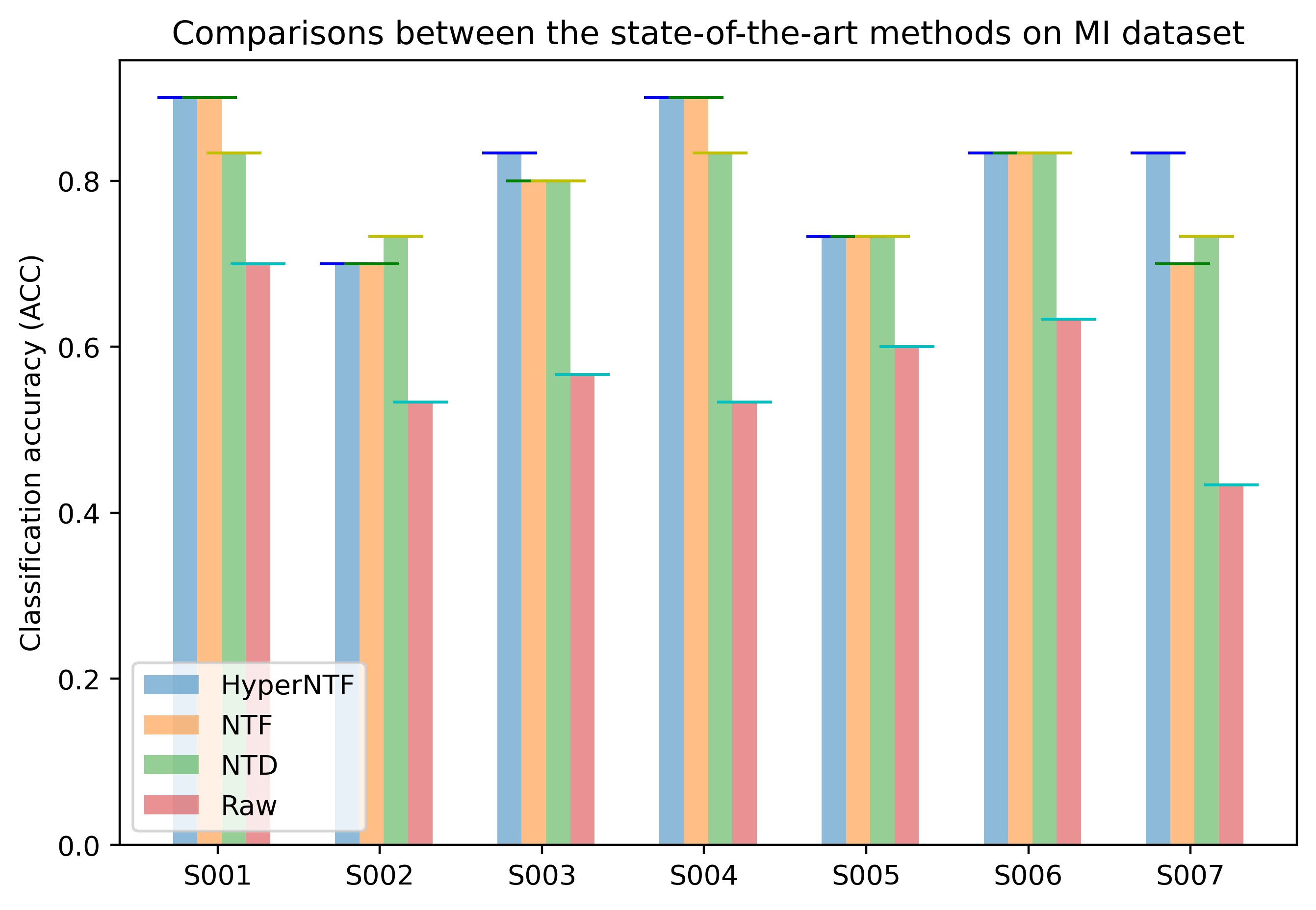} 
      \caption{The average and standard deviations of classification accuracy using LDA classifier by HyperNTF, NTF and NTD algorithms on the EEG MI dataset.}
      \label{MI_classification} 
\end{figure}

\begin{figure}[htb]  
      \centering  
      \includegraphics[width=0.95\linewidth]{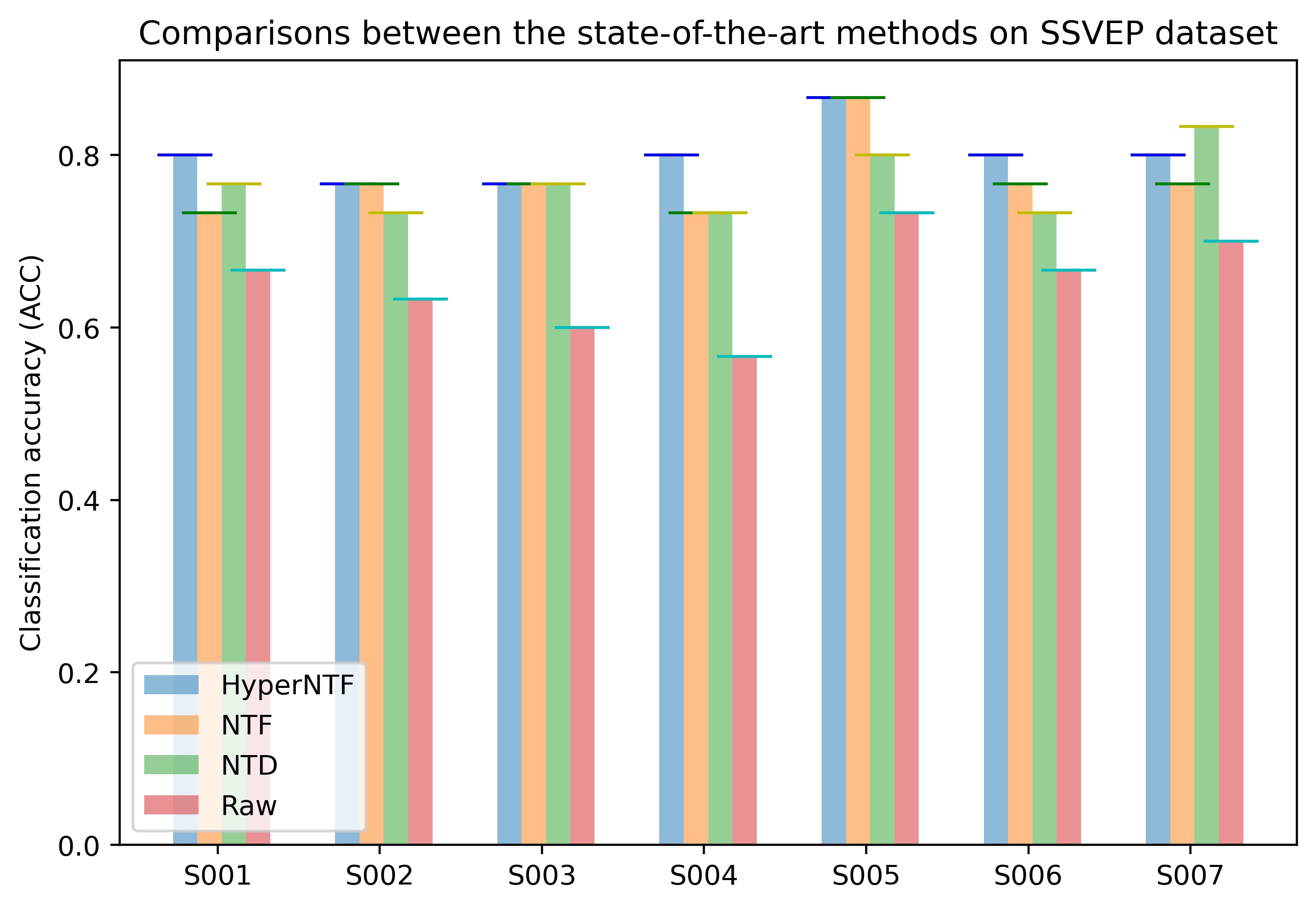} 
      \caption{The average and standard deviations of classification accuracy using LDA classifier by HyperNTF, NTF and NTD algorithms on the SSVEP dataset.}
      \label{SSVEP_classification} 
\end{figure}

In tensor-based EEG classification task, we firstly obtain the reduced data by using HyperNTF, NTF and NTD, and then input the extracted features, as well as the raw data, into a standard linear discriminant analysis (LDA) classifier. We use 5-fold-cross validation to obtain the averaged classification accuracy in testing samples. \textbf{Figure~\ref{MI_classification}} and \textbf{Figure~\ref{SSVEP_classification}} illustrate the mean and standard deviation of classification accuracy on the MI and SSVEP datasets. 

As shown in \textbf{Figure~\ref{MI_classification}}, in EEG-based MI classification experiments, HyperNTF does not have a reliable performance better than NTD and NTF algorithms, but in comparison to the raw EEG data, it has a significant improvement.

From \textbf{Figure~\ref{SSVEP_classification}} we can see that in comparison to NTD, NTF and raw EEG data for classification, HyperNTF cannot gets a better classification result. Therefore, HyperNTF is more suitable for dealing with natural image data, for revealing the non-linear structures in dimensionality reduction.

\section{Discussions}\label{discussion}
From perspective of practical application, manifold learning and tensor decomposition are two of the most widely-used methods for dimensionality reduction. However, the former are nonlinear feature extraction techniques and usually face the challenge of out-of-sample problem, the latter are multi-linear subspace learning techniques. In order to exploit the advantage of two methods, our proposed Hypergraph Regularized Non-negative Tensor Factorization (HyperNTF), can simultaneously conducts dimensionality reduction and deals with nonlinear structure of natural images and EEG signals, facilitating the downstream tasks of clustering and classification. In HyperNTF, the first $ (N-1) $-modes is conducted dimensionality reduction and last mode is expressed as the low-dimensional representation for input data. 

In comparison to the Tucker-based methods, the number of learning parameters in HyperNTF can be largely reduced. For instance, given a $ N\mathrm{th} $-order tensor, \textit{i.e.} $\mathcal{X}\in \mathbb{R}^{L_1\times L_2\cdots\times M} $, the sum of input parameters is $ M {\textstyle \prod_{n=1}^{N-1}}L_n  $. Totally, it needs to learn $ M\times J{\textstyle \sum_{n=1}^{N-1}}J_n\times L_n $ parameters in Tucker decomposition, but in the case of CP decomposition, it only needs to learn $ {\textstyle \sum_{n=1}^{N-1}}J_n\times L_n $ parameters.

Then, we compare the storage consumption among the state-of-the-art methods. For the low-rank regularized heterogeneous tensor decomposition (LRRHTD), which was proposed for subspace clustering~\cite{zhang2017low}, assuming that the factor matrix corresponding with last mode of Tucker decomposition with low rank constraint and the other modes of factor matrices with orthogonal constraint. Due to the factor matrix ($ Z\in \mathbb{R}^{M\times M} $) associated with last mode of Tucker decomposition is a square matrix, as the number of training samples is increasing, its storage and computational consumption is square increased. In contrast, HyperNTF only needs to $ M\times J $ storage, and $J << M$. For Tucker decomposition, due to the core tensor is utilized as the low-dimensional representation of input data, resulting in the storage consumption of $ M{\textstyle \prod_{n=1}^{N-1}J_n} $ parameters. Hence, in comparison to LRRHTD and Tucker model, HyperNTF has the least of storage consumption.

Moreover, to validate the effectiveness of our proposed algorithm, we conduct the manifold unfolding, image clustering, and EEG classification experiments. The experimental results demonstrate that hypergraph can obtain a reliable reconstruction from $ D=3 $ to $ D=2 $ dimensions if only have an optimal selection of $ k $-nearest neighborhoods (\textbf{Figure 3-6}). All results demonstrate that hypergraph can effectively reveal the complex structures or higher-order similarity within original data. In the cluster analysis (\textbf{Figure 9-12}), HyperNTF is superior to the comparison algorithms, including HTD-Multinomial, LRRHTD, GLTD, and HOSVD, regardless of the different cluster numbers. Therefore, our proposed HyperNTF has a distinct advantage in clustering of image data. However, for EEG classification experiment, HyperNTF has a closely similar performance with NTD and NTF algorithms (\textbf{Figure 11-12}), suggesting that HyperNTF is more suitable for clustering rather than classification. 

\section{Conclusions and Future Work}
In this paper, we develop a new method for dimensionality reduction, named Hypergraph Regularized Non-negative Tensor Factorization (HyperNTF). We conduct a variety of experiments to show the effectiveness of our proposed algorithm, including the unfolding of synthetic manifold, clustering and classification experiments. In our future work, we can carry out in several different directions. For instance, in our current work we use an alternating procedure to solve the factor matrices and low-dimensional representation. However, this is not the only way to solve the objective function. As the constraints $ \boldsymbol{1} U_n=\boldsymbol{1}, U_n\geq 0 $ for $ n=1,\ldots, N-1 $ defines the Multinomial manifold, the overall optimization problem can be defined over the product manifold of $ \left( N-1 \right)$-Multinomial manifolds and the last one on the Euclidean space. In addition, it is worth investigating on how to initialize the factor matrix, how to select the best order of projection, and how to select the best metrics and stopping criteria.

\section*{Acknowledgements}
The authors would like to thank Dr. Haiyan Wu for sharing the EEG data. This work was funded in part by the National Natural Science Foundation of China (62001205), National Key Research and Development Program of China (2021YFF1200800), Guangdong Natural Science Foundation Joint Fund (2019A1515111038), Shenzhen Science and Technology Innovation Committee (20200925155957004, KCXFZ2020122117340001), Shenzhen-Hong Kong-Macao Science and Technology Innovation Project (SGDX2020110309280100), Shenzhen Key Laboratory of Smart Healthcare Engineering (ZDSYS20200811144003009).


\printbibliography

@article{tenenbaum290global,
  title={A global geometric framework for nonlinear dimension reduction},
  author={Tenenbaum, J and De Silva, V and Langford, J},
  journal={Science},
  volume={290}
}

@article{roweis2000nonlinear,
  title={Nonlinear dimensionality reduction by locally linear embedding},
  author={Roweis, Sam T and Saul, Lawrence K},
  journal={science},
  volume={290},
  number={5500},
  pages={2323--2326},
  year={2000},
  publisher={American Association for the Advancement of Science}
}

@article{zhang2004principal,
  title={Principal manifolds and nonlinear dimensionality reduction via tangent space alignment},
  author={Zhang, Zhenyue and Zha, Hongyuan},
  journal={SIAM journal on scientific computing},
  volume={26},
  number={1},
  pages={313--338},
  year={2004},
  publisher={SIAM}
}

@article{belkin2003laplacian,
  title={Laplacian eigenmaps for dimensionality reduction and data representation},
  author={Belkin, Mikhail and Niyogi, Partha},
  journal={Neural computation},
  volume={15},
  number={6},
  pages={1373--1396},
  year={2003},
  publisher={MIT Press}
}

@article{savas2007handwritten,
  title={Handwritten digit classification using higher order singular value decomposition},
  author={Savas, Berkant and Eld{\'e}n, Lars},
  journal={Pattern recognition},
  volume={40},
  number={3},
  pages={993--1003},
  year={2007},
  publisher={Elsevier}
}

@inproceedings{kim2007nonnegative,
  title={Nonnegative tucker decomposition},
  author={Kim, Yong-Deok and Choi, Seungjin},
  booktitle={2007 IEEE Conference on Computer Vision and Pattern Recognition},
  pages={1--8},
  year={2007},
  organization={IEEE}
}

@article{li2016mr,
  title={MR-NTD: Manifold regularization nonnegative tucker decomposition for tensor data dimension reduction and representation},
  author={Li, Xutao and Ng, Michael K and Cong, Gao and Ye, Yunming and Wu, Qingyao},
  journal={IEEE transactions on neural networks and learning systems},
  volume={28},
  number={8},
  pages={1787--1800},
  year={2016},
  publisher={IEEE}
}

@article{sun2015heterogeneous,
  title={Heterogeneous tensor decomposition for clustering via manifold optimization},
  author={Sun, Yanfeng and Gao, Junbin and Hong, Xia and Mishra, Bamdev and Yin, Baocai},
  journal={IEEE transactions on pattern analysis and machine intelligence},
  volume={38},
  number={3},
  pages={476--489},
  year={2015},
  publisher={IEEE}
}

@article{zhang2017low,
  title={Low-rank regularized heterogeneous tensor decomposition for subspace clustering},
  author={Zhang, Jing and Li, Xinhui and Jing, Peiguang and Liu, Jing and Su, Yuting},
  journal={IEEE Signal Processing Letters},
  volume={25},
  number={3},
  pages={333--337},
  year={2017},
  publisher={IEEE}
}

@article{hu2014eigenvectors,
  title={The eigenvectors associated with the zero eigenvalues of the Laplacian and signless Laplacian tensors of a uniform hypergraph},
  author={Hu, Shenglong and Qi, Liqun},
  journal={Discrete Applied Mathematics},
  volume={169},
  pages={140--151},
  year={2014},
  publisher={Elsevier}
}

@article{hu2015laplacian,
  title={The Laplacian of a uniform hypergraph},
  author={Hu, Shenglong and Qi, Liqun},
  journal={Journal of Combinatorial Optimization},
  volume={29},
  number={2},
  pages={331--366},
  year={2015},
  publisher={Springer}
}

@article{tian2009hypergraph,
  title={A hypergraph-based learning algorithm for classifying gene expression and arrayCGH data with prior knowledge},
  author={Tian, Ze and Hwang, TaeHyun and Kuang, Rui},
  journal={Bioinformatics},
  volume={25},
  number={21},
  pages={2831--2838},
  year={2009},
  publisher={Oxford University Press}
}

@inproceedings{sun2008hypergraph,
  title={Hypergraph spectral learning for multi-label classification},
  author={Sun, Liang and Ji, Shuiwang and Ye, Jieping},
  booktitle={Proceedings of the 14th ACM SIGKDD international conference on Knowledge discovery and data mining},
  pages={668--676},
  year={2008}
}

@inproceedings{zhou2007learning,
  title={Learning with hypergraphs: Clustering, classification, and embedding},
  author={Zhou, Dengyong and Huang, Jiayuan and Sch{\"o}lkopf, Bernhard},
  booktitle={Advances in neural information processing systems},
  pages={1601--1608},
  year={2007}
}

@article{cichocki2007nonnegative,
  title={Nonnegative matrix and tensor factorization [lecture notes]},
  author={Cichocki, Andrzej and Zdunek, Rafal and Amari, Shun-ichi},
  journal={IEEE signal processing magazine},
  volume={25},
  number={1},
  pages={142--145},
  year={2007},
  publisher={IEEE}
}

@phdthesis{cohen2016environmental,
  title={Environmental multiway data mining},
  author={Cohen, J{\'e}r{\'e}my},
  year={2016}
}

@article{de2000multilinear,
  title={A multilinear singular value decomposition},
  author={De Lathauwer, Lieven and De Moor, Bart and Vandewalle, Joos},
  journal={SIAM journal on Matrix Analysis and Applications},
  volume={21},
  number={4},
  pages={1253--1278},
  year={2000},
  publisher={SIAM}
}

@misc{petersen2012matrix,
  title={The matrix cookbook (version: November 15, 2012)},
  author={Petersen, Kaare Brandt and Pedersen, Michael Syskind},
  year={2012},
  publisher={Technical University of Denmark)(www. math. uwaterloo. ca/\~{} hwolkowi~…}
}

@phdthesis{kaya2017high,
  title={High performance parallel algorithms for tensor decompositions},
  author={Kaya, Oguz},
  year={2017}
}

@article{jiang2018image,
  title={Image representation and learning with graph-laplacian tucker tensor decomposition},
  author={Jiang, Bo and Ding, Chris and Tang, Jin and Luo, Bin},
  journal={IEEE transactions on cybernetics},
  volume={49},
  number={4},
  pages={1417--1426},
  year={2018},
  publisher={IEEE}
}

@article{lee1999learning,
  title={Learning the parts of objects by non-negative matrix factorization},
  author={Lee, Daniel D and Seung, H Sebastian},
  journal={Nature},
  volume={401},
  number={6755},
  pages={788--791},
  year={1999},
  publisher={Nature Publishing Group}
}

@inproceedings{lee2001algorithms,
  title={Algorithms for non-negative matrix factorization},
  author={Lee, Daniel D and Seung, H Sebastian},
  booktitle={Advances in neural information processing systems},
  pages={556--562},
  year={2001}
}

@article{qiao2012explicit,
  title={An explicit nonlinear mapping for manifold learning},
  author={Qiao, Hong and Zhang, Peng and Wang, Di and Zhang, Bo},
  journal={IEEE transactions on cybernetics},
  volume={43},
  number={1},
  pages={51--63},
  year={2012},
  publisher={IEEE}
}

@inproceedings{cai2007learning,
  title={Learning a spatially smooth subspace for face recognition},
  author={Cai, Deng and He, Xiaofei and Hu, Yuxiao and Han, Jiawei and Huang, Thomas},
  booktitle={2007 IEEE Conference on Computer Vision and Pattern Recognition},
  pages={1--7},
  year={2007},
  organization={IEEE}
}

@article{cruceru2020computationally,
  title={Computationally Tractable Riemannian Manifolds for Graph Embeddings},
  author={Cruceru, Calin and B{\'e}cigneul, Gary and Ganea, Octavian-Eugen},
  journal={arXiv preprint arXiv:2002.08665},
  year={2020}
}

@article{wang2011image,
  title={Image representation using Laplacian regularized nonnegative tensor factorization},
  author={Wang, Can and He, Xiaofei and Bu, Jiajun and Chen, Zhengguang and Chen, Chun and Guan, Ziyu},
  journal={Pattern Recognition},
  volume={44},
  number={10-11},
  pages={2516--2526},
  year={2011},
  publisher={Elsevier}
}

@inproceedings{wang2006tensor,
  title={Tensor discriminant analysis for view-based object recognition},
  author={Wang, Yong and Gong, Shaogang},
  booktitle={18th International Conference on Pattern Recognition (ICPR'06)},
  volume={3},
  pages={33--36},
  year={2006},
  organization={IEEE}
}

@inproceedings{tao2006elapsed,
  title={Elapsed time in human gait recognition: A new approach},
  author={Tao, Dacheng and Li, Xuelong and Wu, Xindong and Maybank, Steve},
  booktitle={2006 IEEE International Conference on Acoustics Speech and Signal Processing Proceedings},
  volume={2},
  pages={II--II},
  year={2006},
  organization={IEEE}
}

@inproceedings{hua2007face,
  title={Face recognition using discriminatively trained orthogonal rank one tensor projections},
  author={Hua, Gang and Viola, Paul A and Drucker, Steven M},
  booktitle={2007 IEEE Conference on Computer Vision and Pattern Recognition},
  pages={1--8},
  year={2007},
  organization={IEEE}
}

@article{liu2017detecting,
  title={Detecting large-scale networks in the human brain using high-density electroencephalography},
  author={Liu, Quanying and Farahibozorg, Seyedehrezvan and Porcaro, Camillo and Wenderoth, Nicole and Mantini, Dante},
  journal={Human brain mapping},
  volume={38},
  number={9},
  pages={4631--4643},
  year={2017},
  publisher={Wiley Online Library}
}

@article{zhao2022fast,
  title={Fast hypergraph regularized nonnegative tensor ring decomposition based on low-rank approximation},
  author={Zhao, Xinhai and Yu, Yuyuan and Zhou, Guoxu and Zhao, Qibin and Sun, Weijun},
  journal={Applied Intelligence},
  pages={1--24},
  year={2022},
  publisher={Springer}
}

@article{wu2020advancing,
  title={Advancing non-negative latent factorization of tensors with diversified regularizations},
  author={Wu, Hao and Luo, Xin and Zhou, MengChu},
  journal={IEEE Transactions on Services Computing},
  year={2020},
  publisher={IEEE}
}

@article{wu2019posterior,
  title={A posterior-neighborhood-regularized latent factor model for highly accurate web service QoS prediction},
  author={Wu, Di and He, Qiang and Luo, Xin and Shang, Mingsheng and He, Yi and Wang, Guoyin},
  journal={IEEE Transactions on Services Computing},
  year={2019},
  publisher={IEEE}
}

@article{liu2020convergence,
  title={Convergence analysis of single latent factor-dependent, nonnegative, and multiplicative update-based nonnegative latent factor models},
  author={Liu, Zhigang and Luo, Xin and Wang, Zidong},
  journal={IEEE Transactions on Neural Networks and Learning Systems},
  volume={32},
  number={4},
  pages={1737--1749},
  year={2020},
  publisher={IEEE}
}

@article{balasubramaniam2020column,
  title={Column-wise element selection for computationally efficient nonnegative coupled matrix tensor factorization},
  author={Balasubramaniam, Thirunavukarasu and Nayak, Richi and Yuen, Chau and Tian, Yu-Chu},
  journal={IEEE Transactions on Knowledge and Data Engineering},
  volume={33},
  number={9},
  pages={3173--3186},
  year={2020},
  publisher={IEEE}
}

@article{jokinen2019clustering,
  title={Clustering structure analysis in time-series data with density-based clusterability measure},
  author={Jokinen, Juho and R{\"a}ty, Tomi and Lintonen, Timo},
  journal={IEEE/CAA Journal of Automatica Sinica},
  volume={6},
  number={6},
  pages={1332--1343},
  year={2019},
  publisher={IEEE}
}

@article{liu2019embedded,
  title={An embedded feature selection method for imbalanced data classification},
  author={Liu, Haoyue and Zhou, MengChu and Liu, Qing},
  journal={IEEE/CAA Journal of Automatica Sinica},
  volume={6},
  number={3},
  pages={703--715},
  year={2019},
  publisher={IEEE}
}

@article{lu2018structurally,
  title={Structurally incoherent low-rank nonnegative matrix factorization for image classification},
  author={Lu, Yuwu and Yuan, Chun and Zhu, Wenwu and Li, Xuelong},
  journal={IEEE Transactions on Image Processing},
  volume={27},
  number={11},
  pages={5248--5260},
  year={2018},
  publisher={IEEE}
}

@article{huang2018improved,
  title={Improved hypergraph regularized Nonnegative Matrix Factorization with sparse representation},
  author={Huang, Sheng and Wang, Hongxing and Ge, Yongxin and Huangfu, Luwen and Zhang, Xiaohong and Yang, Dan},
  journal={Pattern Recognition Letters},
  volume={102},
  pages={8--14},
  year={2018},
  publisher={Elsevier}
}

@article{lu2009regularized,
  title={Regularized locality preserving projections and its extensions for face recognition},
  author={Lu, Jiwen and Tan, Yap-Peng},
  journal={IEEE Transactions on Systems, Man, and Cybernetics, Part B (Cybernetics)},
  volume={40},
  number={3},
  pages={958--963},
  year={2009},
  publisher={IEEE}
}

@article{lu2015low,
  title={Low-rank preserving projections},
  author={Lu, Yuwu and Lai, Zhihui and Xu, Yong and Li, Xuelong and Zhang, David and Yuan, Chun},
  journal={IEEE transactions on cybernetics},
  volume={46},
  number={8},
  pages={1900--1913},
  year={2015},
  publisher={IEEE}
}

@article{schalk2004bci2000,
  title={BCI2000: a general-purpose brain-computer interface (BCI) system},
  author={Schalk, Gerwin and McFarland, Dennis J and Hinterberger, Thilo and Birbaumer, Niels and Wolpaw, Jonathan R},
  journal={IEEE Transactions on biomedical engineering},
  volume={51},
  number={6},
  pages={1034--1043},
  year={2004},
  publisher={IEEE}
}

\end{document}